\documentclass{article}

\usepackage[preprint]{neurips_2021}

\usepackage[utf8]{inputenc} 
\usepackage[T1]{fontenc}    
\usepackage{hyperref}       
\usepackage{url}            
\usepackage{booktabs}       
\usepackage{amsfonts}       
\usepackage{nicefrac}       
\usepackage{microtype}      
\usepackage{xcolor}         

\usepackage{bm}
\usepackage{microtype}
\usepackage{graphicx}
\usepackage{subcaption}
\usepackage{booktabs} 
\usepackage{amssymb}
\usepackage{amsmath}

\usepackage{algorithm}
\usepackage{algorithmic}

\usepackage{todonotes}

\usepackage{hyperref}
\usepackage{cleveref}
\usepackage{microtype}

\usepackage{import}

\newcommand{\norm}[1]{\left\lVert #1 \right\rVert}
\newcommand{\floor}[1]{\lfloor #1 \rfloor}

\DeclareMathOperator{\dist}{dist}

\DeclareMathOperator{\cossim}{sim}

\usepackage[acronym,numberedsection]{glossaries}
\makeglossaries
\newacronym{ad}{AD}{Anomaly Detection}
\newacronym{arima}{ARIMA}{Autoregressive Integrated Moving Average}
\newacronym{bce}{BCE}{Binary Cross-Entropy}
\newacronym{cnn}{CNN}{Convolutional Neural Network}
\newacronym{coe}{COE}{Contextual Outlier Exposure}
\newacronym{gan}{GAN}{Generative Adversarial Network}
\newacronym{hsc}{HSC}{Hypersphere Classifier}
\newacronym{mae}{MAE}{Mean Absolute Error}
\newacronym{mse}{MSE}{Mean Squared Error}
\newacronym{msl}{MSL}{Mars Science Laboratory rover}
\newacronym{ncad}{NCAD}{Neural Contextual Anomaly Detection}
\newacronym{oe}{OE}{Outlier Exposure}
\newacronym{po}{po}{Point Outliers}
\newacronym{smap}{SMAP}{Soil Moisture Active Passive satellite}
\newacronym{smd}{SMD}{Server Machine Dataset}
\newacronym{sgd}{SGD}{Stochastic Gradient Descent}
\newacronym{svm}{SVM}{Support Vector Machine}
\newacronym{swat}{SWaT}{Secure Water Treatment}
\newacronym{tcn}{TCN}{Temporal Convolutional Network}
\newacronym{vae}{VAE}{Variational Auto-Encoder}

\title{Neural Contextual Anomaly Detection for Time Series}

\newcommand*\samethanks[1][\value{footnote}]{\footnotemark[#1]}

\author{%
  Chris U. Carmona
  \thanks{Equal contribution.} \
  \thanks{Work done while working at AWS AI Labs.}
  \\
  University of Oxford\\
  \texttt{carmona@stats.ox.ac.uk}\\
 \And
 Fran\c{c}ois-Xavier Aubet \samethanks[1]
 \\
 AWS AI Labs\\
 \texttt{aubetf@amazon.com}\\
 \And
 Valentin Flunkert\\
 AWS AI Labs\\
 \texttt{flunkert@amazon.com}\\
 \And
 Jan Gasthaus\\
 AWS AI Labs\\
 \texttt{gasthaus@amazon.com}
}

\begin{document}

\maketitle

\begin{abstract}

We introduce Neural Contextual Anomaly Detection (NCAD), a framework for anomaly detection on time series that scales seamlessly from the unsupervised to supervised setting, and is applicable to both univariate and multivariate time series.
This is achieved by effectively combining recent developments in representation learning for multivariate time series, with techniques for deep anomaly detection originally developed for computer vision that we tailor to the time series setting.
Our window-based approach facilitates learning the boundary between normal and anomalous classes by injecting generic synthetic anomalies into the available data.
Moreover, our method can effectively take advantage of all the available information, be it as domain knowledge, or as training labels in the semi-supervised setting.
We demonstrate empirically on standard benchmark datasets that our approach obtains a state-of-the-art performance in these settings.

\end{abstract}

\vspace{-5pt}
\section{Introduction}
\label{sec:introduction}
Detecting anomalies in real-valued time series data has many practical applications, such as monitoring machinery for faults, finding anomalous behavior in IoT sensor data, improving the availability of computer applications and (cloud) infrastructure, and monitoring patients vital signs, among many others. Since Shewhart's pioneering work on statistical process control \citep{shewhart1931economic}, statistical techniques for monitoring and detecting abnormal behavior have been developed, refined, and deployed in countless highly impactful applications.

Recently, deep learning techniques have been successfully applied to various anomaly detection problems (see e.g.\ the surveys by \citet{Ruff2020adreview} and \citet{pang2020deep}).
In the particular case of time series, these methods have demonstrated remarkable performance for large-scale monitoring problems such as those encountered by companies like Google \citep{shipmon2017time}, Microsoft \citep{Ren2019srcnn}, Alibaba \citep{Gao2020robustad}, and Amazon \citep{ayed2020anomaly}.


Classically, anomaly detection on time series is cast as an unsupervised learning problem, where the training data contains both normal and anomalous instances, but without knowing which is which. However, in many practical applications, a \emph{fully} unsupervised approach can leave valuable information unutilized, as it is often possible to obtain (small amounts of) labeled anomalous instances, or to characterize the relevant anomalies in some general way.
%

Ideally, an effective method for anomaly detection requires a semi-supervised approach, allowing to utilize information about known anomalous patterns or out-of-distribution observations, if any of these are available. Recent developments on deep anomaly detection for computer vision have achieved remarkable performance by following such a learning strategy. A notable example is the line of work leading to the \acrlong{hsc} \citep{Ruff2018deepsvd, Ruff2020deepsad, Ruff2020hsc}, which extends the concept of one-class classification to a powerful framework for semi-supervised anomaly detection on complex data.

%

%
%


In this work, we introduce \acrfull{ncad}, a framework for anomaly detection on time series that can scale seamlessly from the unsupervised to supervised setting, allowing to incorporate additional information, both through labeled examples and through known anomalous patterns.
This is achieved by effectively combining recent developments in representation learning for multivariate time series \citep{Franceschi2019}, with a number of deep anomaly detection techniques originally developed for computer vision, such as the \acrfull{hsc} \citep{Ruff2020hsc} and \acrfull{oe} \citep{Hendrycks2019oe}, but tailored to the time series setting.

Our approach is based on breaking each time series into overlapping, fixed-size windows. Each window is further divided into two parts: a \emph{context window} and a \emph{suspect window} (see \cref{fig:ncad_visualisation}), which are mapped into neural representations (embedding) using \acrfullpl{tcn} \citep{Bai2018tcn}.
Our aim is to detect anomalies in the \emph{suspect window}. Anomalies are identified in the space of learned latent representations, building on the intuition that anomalies create a substantial perturbation on the embeddings, so when we compare the representation of two overlapping segments, one with the anomaly and one without it, we expect them to be distant.

Time series anomalies are inherently contextual. We account for this in our methodology by extending the \acrshort{hsc} loss to a \emph{contextual} hypersphere loss, which dynamically adapts the hypersphere's center based on the context's representation.
%
We use data augmentation techniques to ease the learning of the boundary between the normal and anomalous classes. In particular, we employ a variant of OE to create contextual anomalies, and employ simple injected point outlier anomalies.

In summary, we make the following contributions: 
\textbf{(I)} Propose a simple yet effective framework for time series anomaly detection that achieves state-of-the-art performance across well-known benchmark datasets, covering univariate and multivariate time series, and across the unsupervised,  semi-supervised, and fully-supervised settings (Our implementation of \acrshort{ncad} is publicly available \footnote{\url{https://github.com/Francois-Aubet/gluon-ts/tree/adding_ncad_to_nursery/src/gluonts/nursery/ncad}});
\textbf{(II)} Build on related work on deep anomaly detection using the hypersphere classifier \citep{Ruff2020hsc} and expand it to introduce contextual hypersphere detection.
\textbf{(III)}  Adapt the \acrlong{oe} \citep{Hendrycks2019oe} and Mixup \citep{Zhang2018mixup} methods to the particular case of anomaly detection for time series.




\vspace{-5pt}
\section{Related work}
\label{sec:related_work}

\acrfull{ad} is an important problem with many applications and has consequently been widely studied. We refer the reader to one of the recent reviews in the topic for a general overview of methods \citep{anom4,Ruff2020adreview,pang2020deep}.

We are interested in anomaly detection for time series. This is a problem typically framed in an unsupervised way. A traditional approach is to use a predictive model, estimating the distribution (or confidence bands) of future values conditioned on historical observations, and mark observations as anomalous if they are considered unlikely under the model \citep{shipmon2017time}. Forecasting models such as \acrshort{arima} or exponential smoothing methods are often used here, assuming a Gaussian noise distribution. \citet{dspot} propose \textsc{SPOT} and \textsc{DSPOT}, which detect outliers in time series using extreme value theory to model the tail of the distribution.


New advances on deep learning models for anomaly detection have become popular recently. For time series, some models have maintained the classical predictive approach, and introduced flexible neural networks for the dependency structure, yielding significant improvements. \citet{shipmon2017time} use deep (recurrent) neural networks to parametrize a Gaussian distribution and use the tail probability to detect outliers.

Effective ideas for deep anomaly detection that deviate from the predictive approach have been successfully imported to the time series domain from other fields. 
Reconstruction based methods, e.g. with \acrfullpl{vae}, or density based methods , e.g. with \acrfullpl{gan}: \textsc{Donut} uses a \acrshort{vae} to predict the distribution of sliding windows; \textsc{LSTM-VAE} \citep{lstmvae} uses a recurrent neural network with a \acrshort{vae}; \textsc{OmniAnomaly} \citep{omnianomaly} extends this framework with deep innovation state space models and normalizing flows; \textsc{AnoGAN} \citep{anogan} uses GANs to model sequences of observations and estimate their probabilities in latent space.
\textsc{MSCRED} \citep{Zhang2019mscred} uses convolutional auto-encoders and identifies anomalies by measuring the reconstruction error.

Compression-based approaches have become very popular in image anomaly detection.
The working principle is similar to the one-class classification used in the support vector data description method \citep{svdd}: instances are mapped to latent representations which are pulled together during training, forming a sphere in the latent space; instances that are distant from the center are considered anomalous. \citep{Ruff2018deepsvd, Ruff2020deepsad} build on this idea to learn a neural mapping $\phi( \cdot; \theta)  : \mathbb{R}^D \rightarrow \mathbb{R}^E$, such that the representations of nominal points concentrate around a (fixed) center $\bm{c}$, while anomalous points are mapped away from that center. In the unsupervised case, DeepSVDD \citep{Ruff2018deepsvd} achieves this by minimizing the Euclidean distance $\sum_i ||\phi( \bm{w}_i; \theta) - \bm{c}||^2$, subject to a suitable regularization of the mapping and assuming that anomalies are rare.

\textsc{THOC} \citep{thoc} applies this principle to the context of time series,  by extending the model to consider multiple spheres to obtain more convenient representations. 
This method differs from our work in two ways: it relies on a dilated recurrent neural network with skip connections to handle the contextual aspect of the data, we use a much simpler network and use a context window to handle the contextuality.
%
Then, our method can seamlessly handle the semi-supervised setting and benefits from our data augmentation techniques.

\citet{Ruff2020hsc} propose \acrfull{hsc}, improving on DeepSVDD by training the network using the standard \acrfull{bce} loss, this way extending the approach to the (semi-)supervised setting. With this method, they can rely  on labeled examples to regularize the training and do not have to resort to limiting the network.
In particular, the \acrshort{hsc} loss is given by setting the pseudo-probability of an anomalous instance ($y=1$) as $p=1-\ell(\phi(\bm{w}_i ; \theta))$, i.e.
\begin{equation}\label{eq:hsc_loss}
    -  (1-y_i) \log \ell(\phi(\bm{w}_i ; \theta)) - y_i \log(1 - \ell(\phi(\bm{w}_i ; \theta))) \;,
\end{equation}
where $\ell : \mathbb{R}^E \rightarrow [0,1]$ maps the representation to a probabilistic prediction.
Choosing $\ell(\bm{z}) = \exp(-||\bm{z}||^2)$, leads to a spherical decision boundary in representation space, and reduces to the DeepSVDD loss (with center $\bm{c}=0$) when all labels are $0$. 

%
%

Current work on semi-supervised anomaly detection indicates that including even only few labeled anomalies can already yield remarkable performance improvements on complex data \citep{Ruff2020adreview, Liznerski2020explainablead, Tuluptceva2020ad}. A powerful resource in this line is \acrfull{oe} \citep{Hendrycks2019oe}, which improves detection by incorporating large amounts of out-of-distribution examples from auxiliary datasets during training. Despite such negative samples may not coincide with ground-truth anomalies, such contrasting can be beneficial for learning characteristic representations of normal concepts. Moreover, the combination of \acrlong{oe} and expressive representations with the hyperphere classifier have shown exceptional results for Deep \acrshort{ad} on images \citep{Ruff2020hsc, Deecke2020residual}.

For time series data, however, artificial anomalies and related data augmentation techniques have not been studied extensively. 
\citet{smolyakovLearningEnsemblesAnomaly2019} used artificial anomalies to select thresholds in ensembles of anomaly detection models.
Most closely related to our approach, \textsc{SR-CNN} \cite{Ren2019srcnn} trains a supervised CNN on top of an unsupervised anomaly detection model (\textsc{SR}), by using labels from injected single point outliers. 

Fully supervised methods are not as widely studied because labeling all the anomalies is too expensive and unreliable in most applications. 
An exception is the work by \citet{liu2015opprentice} who propose a system to continuously collect anomaly labels and to iteratively re-train and deploy a supervised random forest model. The \textsc{U-Net-DeWA} approach of \citet{Gao2020robustad} relies on preprocessing using robust time series decomposition to train a convolutional network in a supervised way, relying on data augmentations that preserve the anomaly labels to increase the training set size.

\vspace{-5pt}
\section{Neural Contextual Anomaly Detection}
\label{sec:methods}

This section describes our anomaly detection framework and its building blocks. We combine a window-based anomaly detection approach with a flexible training paradigm 
and effective heuristics for data augmentation to produce a state-of-the-art system for anomaly detection.


We consider the following general time series anomaly detection problem: 
We are given a collection of $N$ discrete-time time series
$\bm{x}^{(i)}_{1:T_i}$, $i=1,\ldots,N$ where for time series $i$ and time step $t=1, \ldots, T_i$ we have an observation vector $\bm{x}_t^{(i)} \in \mathbb{R}^D$.
We further assume that we are given a corresponding, set of partial anomaly labels
$y^{(i)}_{1:T_i}$ with $y^{(i)}_t \in \{0, 1, ?\}$, indicating whether the corresponding 
observation $\bm{x}^{(i)}_t$ is normal ($0$), anomalous ($1$), or unlabeled ($?$).

The goal is to predict anomaly labels $\hat{y}_{1:T}$, with $y_t \in \{0, 1\}$ given a time series $\bm{x}_{1:T}$.
Instead of predicting the binary labels directly, we predict a positive anomaly score for each time step, which can subsequently be thresholded to obtain anomaly labels satisfying a desired precision/recall trade-off.

\subsection{Window-based Contextual Hypersphere Detection}
\label{subsec:contextual_hypersphere}

Similar to other work on time series \acrshort{ad} (e.g.\ \citet{Ren2019srcnn, rcf}),
we convert the time series problem 
to a vector problem by splitting each time series into a sequence of overlapping, fixed-size windows $\bm{w}$ of length $L$. 
A key element of our approach is that within each window, we identify two segments: a \emph{context window} $\bm{w}^{(c)}$ of length $C$ and \emph{suspect window} of length $S$: $\bm{w}=(\bm{w}^{(c)}, \bm{w}^{(s)})$, where we typically choose $C\gg S$. 
Our goal is to detect anomalies in the suspect window relative to the local context provided by the context window.
This split not only naturally aligns with the typically contextual nature of anomalies in time series data, it also allows for short suspect windows (even $S=1$), minimizing detection delays and improving anomaly localization.

Intuitively, our approach is based on the idea that we can identify anomalies by comparing representation vectors $\phi(\bm{w}; \theta)$ and $\phi(\bm{w}^{(c)}; \theta)$, obtained by applying a neural network feature extractor $\phi(\cdot; \theta)$, which is trained in such a way that representations are pulled together if there is no anomaly present in the suspect window $\bm{w}^{(s)}$, and pushed apart otherwise. 

%

We propose a loss function, which can be seen as \emph{contextual} version of the \acrlong{hsc} (equation \ref{eq:hsc_loss}) by considering a loss function which contrasts
the representation of the context window with the representation of the full window:
\begin{equation*}\label{eq:our_generic_loss}
    - (1-y_i) \log\left( \ell\left( \dist(\phi(\bm{w}_i; \theta), \phi(\bm{w}^{(c)}_i; \theta) \right) \right) - y_i \log\left(1 - \ell \left( \dist(\phi(\bm{w}_i; \theta), \phi(\bm{w}^{(c)}_i; \theta)\right)\right) \;.
\end{equation*}
In our experiments we follow \cite{Ruff2020hsc} and use the Euclidean distance $\dist(x,z) = \norm{x - z}_{2}$ and a radial basis function $\ell(z) := \exp (-z^2)$, to create a spherical decision boundary as in \acrshort{hsc}/DeepSVDD, resulting in the loss function
\begin{equation}\label{eq:our_l2_loss}
    (1-y_i) \norm{ \phi(\bm{w}_i; \theta) - \phi(\bm{w}^{(c)}_i; \theta)}_{2}^2 - y_i \log\left( 1 - \exp \left( -\norm{\phi(\bm{w}_i; \theta) - \phi(\bm{w}^{(c)}_i; \theta)}_{2}^2 \right) \right) \;.
\end{equation}
Intuitively, this is the \acrshort{hsc} loss where the center $\bm{c}$ of the  hypersphere is chosen dynamically for each instance as the representation of the context. This introduces an inductive bias: representations of the context window and representations of the full window should be different if an anomaly occurs in the suspect window.
As we show in our empirical analysis, this inductive bias makes the model more label efficient leads to better generalization.
In particular, we show that when this model is trained using generic injected anomalies such as point outliers, it is able to generalize to the more complex anomalies found in real world datasets.

\subsection{NCAD architecture \& training}
\label{sec:ncad}


Our model identifies anomalies in a space of learned latent representations, building on the intuition that: if an anomaly is present in the suspect window $\bm{w}^{(s)}$, then representation vectors constructed from $\bm{w}$ and $\bm{w}^{(c)}$ should be distant. 

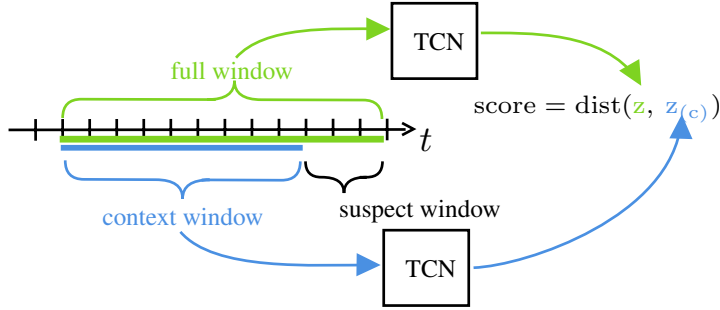
\begin{figure}[tbh]
    \centering
    \resizebox{0.7\columnwidth}{!}{\tikzset{every picture/.style={line width=0.75pt}} 

\begin{tikzpicture}[x=0.75pt,y=0.75pt,yscale=-1,xscale=1]

\draw    (0.72,57.33) -- (147.54,57.17) (10.72,53.32) -- (10.73,61.32)(20.72,53.31) -- (20.73,61.31)(30.72,53.3) -- (30.73,61.3)(40.72,53.29) -- (40.73,61.29)(50.72,53.28) -- (50.72,61.28)(60.72,53.27) -- (60.72,61.27)(70.72,53.26) -- (70.72,61.26)(80.72,53.25) -- (80.72,61.25)(90.72,53.23) -- (90.72,61.23)(100.72,53.22) -- (100.72,61.22)(110.72,53.21) -- (110.72,61.21)(120.72,53.2) -- (120.72,61.2)(130.72,53.19) -- (130.72,61.19)(140.72,53.18) -- (140.72,61.18) ;
\draw   (146.24,54.77) -- (150.48,57.01) -- (146.24,59.25) ;
\draw  [color={rgb, 255:red, 74; green, 144; blue, 226 }  ,draw opacity=1 ] (21.68,68.6) .. controls (21.66,73.27) and (23.98,75.61) .. (28.65,75.63) -- (55.11,75.75) .. controls (61.78,75.78) and (65.1,78.12) .. (65.08,82.79) .. controls (65.1,78.12) and (68.44,75.8) .. (75.11,75.83)(72.11,75.82) -- (101.57,75.95) .. controls (106.24,75.97) and (108.58,73.65) .. (108.61,68.98) ;
\draw  [color={rgb, 255:red, 74; green, 144; blue, 226 }  ,draw opacity=1 ][fill={rgb, 255:red, 74; green, 144; blue, 226 }  ,fill opacity=1 ] (20.53,63.38) -- (108.99,63.38) -- (108.99,64.5) -- (20.53,64.5) -- cycle ;
\draw  [color={rgb, 255:red, 126; green, 211; blue, 33 }  ,draw opacity=1 ][fill={rgb, 255:red, 126; green, 211; blue, 33 }  ,fill opacity=1 ] (20.14,60.02) -- (138.99,60.02) -- (138.99,61.51) -- (20.14,61.51) -- cycle ;
\draw   (110.53,68.97) .. controls (110.53,72.88) and (112.48,74.83) .. (116.39,74.83) -- (116.39,74.83) .. controls (121.97,74.83) and (124.76,76.78) .. (124.76,80.69) .. controls (124.76,76.78) and (127.55,74.83) .. (133.13,74.83)(130.62,74.83) -- (133.13,74.83) .. controls (137.04,74.83) and (138.99,72.88) .. (138.99,68.97) ;
\draw  [color={rgb, 255:red, 126; green, 211; blue, 33 }  ,draw opacity=1 ] (138.8,53.3) .. controls (138.81,48.63) and (136.49,46.29) .. (131.82,46.28) -- (94.24,46.16) .. controls (87.57,46.14) and (84.25,43.8) .. (84.26,39.13) .. controls (84.25,43.8) and (80.91,46.12) .. (74.24,46.1)(77.24,46.11) -- (27.74,45.96) .. controls (23.07,45.94) and (20.73,48.26) .. (20.72,52.93) ;
\draw [color={rgb, 255:red, 74; green, 144; blue, 226 }  ,draw opacity=1 ]   (64.27,95.05) .. controls (76.81,109.72) and (113.11,107.28) .. (135.6,107.31) ;
\draw [shift={(138.33,107.33)}, rotate = 180.75] [fill={rgb, 255:red, 74; green, 144; blue, 226 }  ,fill opacity=1 ][line width=0.08]  [draw opacity=0] (8.93,-4.29) -- (0,0) -- (8.93,4.29) -- cycle    ;
\draw [color={rgb, 255:red, 126; green, 211; blue, 33 }  ,draw opacity=1 ][fill={rgb, 255:red, 255; green, 255; blue, 255 }  ,fill opacity=1 ]   (84.84,30.61) .. controls (95.28,18.82) and (119.98,19.8) .. (138.98,19.98) ;
\draw [shift={(141.94,19.99)}, rotate = 180.16] [fill={rgb, 255:red, 126; green, 211; blue, 33 }  ,fill opacity=1 ][line width=0.08]  [draw opacity=0] (8.93,-4.29) -- (0,0) -- (8.93,4.29) -- cycle    ;
\draw [color={rgb, 255:red, 74; green, 144; blue, 226 }  ,draw opacity=1 ]   (172.67,108.33) .. controls (192.47,107.49) and (245.62,89.59) .. (249.47,54.41) ;
\draw [shift={(249.67,51.67)}, rotate = 452.01] [fill={rgb, 255:red, 74; green, 144; blue, 226 }  ,fill opacity=1 ][line width=0.08]  [draw opacity=0] (8.93,-4.29) -- (0,0) -- (8.93,4.29) -- cycle    ;
\draw [color={rgb, 255:red, 126; green, 211; blue, 33 }  ,draw opacity=1 ][fill={rgb, 255:red, 255; green, 255; blue, 255 }  ,fill opacity=1 ]   (175.38,21.44) .. controls (193.51,21.43) and (220.72,20.73) .. (233.19,38.66) ;
\draw [shift={(234.67,41)}, rotate = 240.15] [fill={rgb, 255:red, 126; green, 211; blue, 33 }  ,fill opacity=1 ][line width=0.08]  [draw opacity=0] (8.93,-4.29) -- (0,0) -- (8.93,4.29) -- cycle    ;
\draw   (142.49,10.46) -- (173.93,10.46) -- (173.93,38.35) -- (142.49,38.35) -- cycle ;

\draw   (139.45,94.44) -- (170.89,94.44) -- (170.89,122.33) -- (139.45,122.33) -- cycle ;

\draw (65.21,84.78) node [anchor=north] [inner sep=0.75pt]  [font=\scriptsize] [align=left] {\textcolor[rgb]{0.29,0.56,0.89}{context window}};
\draw (121.52,81.61) node [anchor=north west][inner sep=0.75pt]  [font=\scriptsize] [align=left] {suspect window};
\draw (83.3,35.29) node  [font=\scriptsize] [align=left] {\textcolor[rgb]{0.49,0.83,0.13}{full window}};
\draw (151.17,55.26) node [anchor=north west][inner sep=0.75pt]   [align=left] {$\displaystyle t$};
\draw (171.23,42.46) node [anchor=north west][inner sep=0.75pt]  [font=\scriptsize] [align=left] {$\displaystyle \mathrm{score=dist(\textcolor[rgb]{0.49,0.83,0.13}{z} ,\ \textcolor[rgb]{0.29,0.56,0.89}{z}\textcolor[rgb]{0.29,0.56,0.89}{_{( c)}})}$};
\draw (149.09,20) node [anchor=north west][inner sep=0.75pt]  [font=\scriptsize] [align=left] {TCN};
\draw (146.05,103.98) node [anchor=north west][inner sep=0.75pt]  [font=\scriptsize] [align=left] {TCN};

\end{tikzpicture}}
    \caption{\acrshort{ncad} encodes two windows that differ by a suspect window using the same TCN network and computes a  distance score of the embeddings. The model is trained to give a high score for instances with an anomaly in the suspect window.
    }
    \label{fig:ncad_visualisation}
\end{figure}


As illustrated in \cref{fig:ncad_visualisation}, our \acrshort{ncad} architecture has three components:\\
\vspace{-10pt}
\begin{enumerate}
\item A neural network \emph{encoder} $\phi( \cdot ; \theta)$, that maps input sequences to representation vectors in $\mathbb{R}^E$. The same encoder is applied both to the full window and to the context window, resulting in representations $z = \phi( \bm{w} ; \theta)$ and $z_{(c)} = \phi( \bm{w}_{(c)} ; \theta)$, respectively.
While any neural network could be used, in our implementation we opt for a \acrshort{cnn} with exponentially dilated causal convolutions \citep{Oord2016wavenet}, in particular the \acrshort{tcn} architecture \citep{Bai2018tcn, Franceschi2019} with adaptive max-pooling along the time dimension.
\item A distance-like function, $\dist( \cdot , \cdot ): \mathbb{R}^{E} \times \mathbb{R}^{E} \rightarrow \mathbb{R}^+$, to compute the similarity between the representations $z$ and $z_{(c)}$.

\item A probabilistic scoring function $\ell(z)=\exp(-\vert z\vert)$, which creates a spherical decision boundary centered at the embedding of the context window.
\end{enumerate}

The parameters $\theta$ of the encoder are learned by minimizing the classification loss on minibatches of windows, $\bm{w}$. These are sampled uniformly at random (across time series and across time) from the training data set $\{\bm{x}^{(i)}_{1:T_i}\}$ after applying the data augmentation techniques that follow.

\paragraph{Rolling predictions} While our window based approach allows the model to decide \emph{if} an anomaly is present in the suspect window, in many applications it is important to react quickly when an anomaly occurs, or to locate the anomaly with some accuracy.
To support these requirements, we apply the model on rolling windows of the time series.
Each time point can then be part of different suspect windows corresponding to different rolling windows had so is given multiple anomaly scores. Using these we can either alert on the first high score, to reduce time to alert, or average the scores for each point to pin-point the anomalies in time more accuratly.

\subsection{Data augmentation}
\label{sec:data_augmentation}


In addition to the contrastive classifier, we utilize a collection of data augmentation methods that inject synthetic anomalies, allowing us to rely on a supervised training objective without requiring ground-truth labels.
While can rely on the hypersphere to train without any labels, having some anomalous examples labels allow to greatly improve the performance, as it has been observed in computer vision \citep{Hendrycks2019oe}. 
We cannot effectively rely on ground-truth anomalies as few datasets have any and in practice it is very costly to obtain training labels; therefore, we propose to generate the anomalous example.
%
%
%
These data augmentation methods explicitly do \emph{not} attempt to characterise the full data distribution of anomalies, which would be infeasible; rather, we combine effective generic heuristics that work well for detecting common types of out-of-distribution examples.

\textbf{\acrfull{coe}} --- Motivated by the success of \acrlong{oe} for \emph{out-of-distribution} detection \cite{Hendrycks2019oe}, we propose a simple task-agnostic method to create contextual out-of-distribution examples. 
Given a data window $\bm{w}=(\bm{w}^{(c)},\bm{w}^{(s)})$, we induce anomalies into the suspect segment, $\bm{w}^{(s)}$, by replacing a chunk of its values with values taken from another time series. 
The replaced values in $\bm{w}^{(s)}$ will most likely break the temporal relation with their neighboring context, therefore creating an out of distribution example.
In our implementation, we apply \acrshort{coe} at training time by selecting random examples in a minibatch and permuting a random length of their suspect windows
(visualizations in \cref{appendix:coe}). In multivariate time series, as anomalies do not have to happen in all dimensions, we randomly select a subset of the dimensions in which the windows are swapped.

\textbf{Anomaly Injection} --- We propose to inject simple single \textbf{\acrfull{po}} in the time series. We use a simple method: at a set of randomly selected time points we add (or subtract) a spike to the time series. The spike is proportional to the inter-quartile range of the points surrounding the spike location.
%
Like for \acrshort{coe}, in multivariate time series we simply select a random subset of dimensions on which we add the spike.
These simple point outliers serve the same purpose as \acrshort{coe}: create clear labeled abnormal points to help the learning of the hypersphere.
(visualizations in \cref{appendix:po}).

In addition to these, in some practical applications, it is possible to identify general characteristics of anomalies that should be detected.
Some widely-known anomalous patterns include: 
sudden changes in the location or scale of the series (change-points); interruption of seasonality, etc.
We have used this approach in our practical application and the domain knowledge allowed to improve the detection performance. As they require and domain knowledge it would be unfair to compare our method when incorporating these; therefore, in the results table we only use the point outliers described above.



\textbf{Window Mixup} --- If we do not have access to training labels and know little about the relevant anomalies, we can only rely on \acrshort{coe} and \acrshort{po}, which may result in significantly missmatch between injected and true anomalies.
To improve generalization of our model in this case, we propose to create linear combinations of training examples inspired by the \textsc{mixup} procedure \cite{Zhang2018mixup}.

\textsc{Mixup} was proposed in the context of computer vision and creates new training examples out of original samples by using a convex combinations of the features and their labels.
This data augmentation technique creates more variety in training examples, but more importantly, the soft labels result in smoother decision functions that generalize better.
\textsc{mixup} is suited for time series applications: 
convex combinations of time series most often result in realistic and plausible new time series (see visualizations in \cref{appendix:mixup}).
We show that  \textsc{mixup} can improve generalization of our model even in cases with a large mismatch between injected and true anomalies.

\vspace{-5pt}
\section{Experiments}
\label{sec:experiments}

In this section, we compare the performance of our approach with alternative methods on public benchmark datasets, and exploring the model behavior under different data settings and model variations in ablation studies.
Further details on the experiments are included in the supplement.

\subsection{Benchmark datasets}
\label{subsec:benchmark_datasets}
We benchmark our method to others on six datasets (more details in \cref{appendix:data_and_assets}):%
\footnote{While we share many of the concerns expressed by \citet{wu2020current} about the lack of quality benchmark datasets for time series anomaly detection, we use these commonly-used benchmark datasets here for lack of better alternatives and to enable direct comparison of our approach to competing methods. 
}

\textbf{\acrfull{smap}} and \textbf{\acrfull{msl}} --- Two datasets published by NASA \cite{Hundman2018telemanon}, with 55 and 27 series respectively. The lengths of the time series vary from 300 to 8500 observations.

\textbf{\acrfull{swat}} --- The dataset was collected on a water treatment testbed over 11 days, 36 attacks were launched in the last 4 days and compose the test set. \citep{mathur2016swat} To compare our numbers with \cite{thoc}, we use the first half of the proposed test set for validation and the second one for test.

\textbf{\acrfull{smd}} --- Is a 5 weeks long dataset with 28 38-dimensional time series each collected from a different machine in large internet companies \citep{omnianomaly}.

\acrshort{smap}, \acrshort{msl}, \acrshort{swat}, and \acrshort{smd}, each have a pre-defined train/test split, where anomalies in the test set are labeled, while the training set contains unlabeled anomalies.

\textbf{\textsc{Yahoo}} --- A dataset published by Yahoo labs,%
\footnote{\url{https://webscope.sandbox.yahoo.com/catalog.php?datatype=s&did=70}}
consisting of 367 real and synthetic time series. 
Following \citep{Ren2019srcnn}, we use the last 50\% of the time points of each of the time series as test set and split the rest in 30\% training and 20\% validation set.

\textbf{\textsc{KPI}} ---  A univariate dataset released in the AIOPS data competition \citep{kpi1}. It consists of KPI curves from different internet companies in 1 minute interval. Like \citep{Ren2019srcnn}, we use 30\% of the train set for validation.
For \textsc{KPI} and \textsc{Yahoo} labels are available for all the series.

\subsection{Evaluation setup}
\vspace{-3pt}
\label{subsec:evaluation_metrics}
Measuring the performance of time series anomaly detection methods in a universal way is challenging, as different applications often require different trade-offs between sensitivity, specificity, and temporal localization.
To account for this, various measures that improve upon simple point-wise classification metrics have been proposed, e.g.\ the flexible segment-based score proposed by \citet{tatbul2018precision} or the score used in the Numenta anomaly benchmark \citep{numenta}.
%
%
To make our results directly comparable, we follow the procedure proposed by \citet{donut} (and subsequently used in other work \cite{omnianomaly, Ren2019srcnn, thoc}), which offers a practical compromise: point-wise scores are used, but the predicted labels are expanded to mark an entire true anomalous segment as detected correctly if at least one time point was detected by the model.%
\footnote{We use the implementation by \citet{omnianomaly}:  \url{https://github.com/NetManAIOps/OmniAnomaly/}.}
We align our experimental protocol with this body of prior work and report $F1$ scores computed by choosing the best threshold on the test set.
For each dataset, the best threshold is chosen and used on all the time series of the test set.

In many real world scenarios, one is interested in detecting anomalies in a streaming fashion on a variety of different time series that may not be known at deployment time. We incorporate these requirements by training a single model on all the training time series of each dataset, and evaluate that model on all the test set time series. 
Further we use short suspect windows allowing to decide if a point is anomalous or not when it is first observed. We report the performance of this harder detection setting.



%

Hyperparameters were chosen in the following way: for \textsc{Yahoo}, \textsc{KPI} and \acrshort{swat}, as the validation datasets have anomaly labels available, we use a Bayesian optimization \citep{perrone2020smtuner} for parameter tuning, by maximizing the F1 score on the validation set.
If no validation set with labels is available, we use a set of standard hyperparameter settings inferred from the datasets with validation datasets. (see details in the supplement).
On each dataset we pick the context window length to roughly match the length of the seasonal patterns in the time series.


We run the model 10 times on each of the benchmark datasets and report mean and standard deviation.
We use standard AWS EC2 ml.p3.2xlarge instances with a single-core Tesla V100 GPU. Training the model on one of the benchmark datasets takes on average 90 minutes.
%
In our code \footnote{\url{https://github.com/Francois-Aubet/gluon-ts/tree/adding_ncad_to_nursery/src/gluonts/nursery/ncad}} we provide scripts to reproduce  the results on the benchmark datasets shown below.

\subsection{Benchmark results}
\label{sec:bmk_results}

\Cref{table:benchmark_univariate} shows the performance of our \acrshort{ncad} approach compared against the state-of-the-art methods on two commonly used univariate datasets.
As these datasets contains labels for anomalies both on the training and the test set, we evaluate our method on them both in the supervised setting (\emph{(sup.)}) and the unsupervised setting (\emph{(un.)})
%
%
We take the numbers from \citet{Ren2019srcnn}.
Our approach significantly outperforms competing approaches on \textsc{Yahoo}, performs similarly to the best unsupervised approach on \textsc{KPI}, and slightly worse than the best supervised approach.
It is important to note that while other methods are either designed for the supervised or unsupervised setting, our method can be used seamlessly in both settings.

\begin{table}[ht]
  \caption{F1 score  of the model on univariate datasets}
  \label{table:benchmark_univariate}
  \centering
  \begin{tabular}{l|cccccccccccc}
        \toprule
        Model & \textsc{Yahoo} (un.) & \textsc{KPI} (un.) & \textsc{KPI} (sup.)  \\  
        \midrule
        SPOT  \citep{dspot}& 33.8 & 21.7 & ---   \\
        DSPOT  \citep{dspot}& 31.6 & 52.1  & ---   \\
        DONUT  \citep{donut}& 2.6 & 34.7 & ---   \\
        SR \citep{Ren2019srcnn} & 56.3 & 62.2 & ---     \\
        SR-CNN  \citep{Ren2019srcnn} & 65.2 & \textbf{77.1} & ---   \\
        SR+DNN  \citep{Ren2019srcnn} & --- & --- & \textbf{81.1}  \\
        \midrule
        \multicolumn{1}{l|}{\acrshort{ncad} w/ \acrshort{coe}, \acrshort{po} , mixup} & \textbf{81.16 $\pm$ 1.43} &  \textbf{76.64 $\pm$ 0.89}  & 79.20 $\pm$ 0.92 \\
        \bottomrule
    \end{tabular}
\end{table}

The \textsc{Yahoo}-supervised experiments are included in \cref{appendix:supervised_yahoo}, where we compare against the supervised approach of \citep{Gao2020robustad}, which represents the state-of-the-art in this setting to the best of our knowledge. 
%
Our approach outperforms their approach significantly with $79\%$ point-wise F1 score versus $69.3\%$ F1 score for their approach.

\begin{table}[ht]
  \caption{F1 score of the model on multivariate datasets}
  \label{table:benchmark_multivariate}
  \centering
  \resizebox{0.99\columnwidth}{!}{%
  \begin{tabular}{lccccccccccccc}
    \toprule
        Model & \acrshort{smap} & \acrshort{msl} & \acrshort{swat} & \acrshort{smd} \\
    \midrule
        AnoGAN \citep{anogan}&  74.59 & 86.39 & 86.64 & --- \\
        DeepSVDD \citep{Ruff2018deepsvd}& 71.71 & 88.12 & 82.82 & --- \\
        DAGMM \citep{dagmm} & 82.04 & 86.08 & 85.38 & 70.94 \\
        LSTM-VAE \citep{lstmvae} & 75.73 & 73.79  & 86.39 & 78.42 \\
        MSCRED \citep{Zhang2019mscred} & 77.45 & 85.97 & 86.84 & --- \\
        OmniAnomaly \citep{omnianomaly} & 84.34 & 89.89 & --- & \textbf{88.57} \\
        MTAD-GAT \citep{zhao2020multivariate} & 90.13 & 90.84 & --- & --- \\ 
        THOC \citep{thoc}&  \textbf{95.18} & 93.67 & 88.09 & --- \\
    \midrule
        \multicolumn{1}{l}{\acrshort{ncad} w/ \acrshort{coe}, \acrshort{po} , mixup}  & \textbf{94.45 $\pm$ 0.68}  &  \textbf{95.60 $\pm$ 0.59} 
        &  \textbf{95.28 $\pm$ 0.76} & 80.16 $\pm$ 0.69 \\
    \bottomrule
    \end{tabular}
    }
\end{table}




\Cref{table:benchmark_multivariate} shows the performance of our \acrshort{ncad} approach compared against the state-of-the-art methods.
None of these datasets provides labels for the anomalies in the training set, all benchmark methods are designed for unsupervised anomaly detection.
%
Our method outperforms \textsc{THOC} by a reasonable margin both on \acrshort{msl} and \acrshort{swat}. On \acrshort{smap} while our average score is slightly lower, the difference is within the variance.
OmniAnomaly \citep{omnianomaly} is the state of the art on \acrshort{smd}, our numbers are only second to theirs. 
We note that OmniAnomaly is considerably more costly and less scalable, since it trains one model for each of the 28 time series of the dataset, while we train a single global model.

\subsection{Ablation study}

\begin{figure*}[ht]
	\begin{subfigure}{.5\columnwidth}
		\caption{Ablation study on \acrshort{smap} and \acrshort{msl}}
          \label{table:ablation_study}
          \centering
          \begin{tabular}{lccccccccccccc}
            \toprule
                Model & \acrshort{smap} & \acrshort{msl} \\
            \midrule
                THOC \citep{thoc}&  95.18 & 93.67 \\
            \midrule
                \multicolumn{1}{l}{\acrshort{ncad} w/ \acrshort{coe}, \acrshort{po} , mixup}  & 94.45 &  95.60 
                \\
                \multicolumn{1}{l}{\hspace{4pt} \small{- \acrshort{po} }}  & 94.28 & 94.73  \\ 
                \multicolumn{1}{l}{\hspace{4pt} \small{- \acrshort{coe}}}  & 88.59 & 94.66   \\
                \multicolumn{1}{l}{\hspace{4pt} \small{- mixup - \acrshort{coe} - \acrshort{po}}}  & 66.9 & 79.47    \\ 
                \multicolumn{1}{l}{\hspace{4pt} \small{- contextual - mixup - \acrshort{coe} - \acrshort{po}}}  & 55.09  & 36.03   \\ 
            \bottomrule
            \end{tabular}
	\end{subfigure}
	\begin{subfigure}{.5\columnwidth}
    	\centering 
        \caption{F1 score of \acrshort{ncad} on the \textsc{Yahoo} dataset trained with only a fraction of training anomalies being labeled.}
        \includegraphics[width=.99\columnwidth]{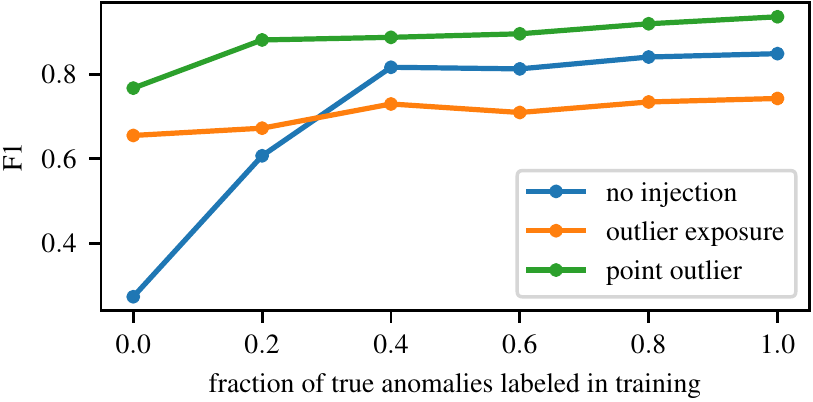}
        \label{fig:percentage_anomalie_v1}
	\end{subfigure}
\label{fig:synthetic_ts}
\end{figure*}

To better understand the advantage brought by each of the components of our method, we perform an ablation study on the \acrshort{smap} and \acrshort{msl} datasets, shown in \cref{table:ablation_study}.
We average two runs for each configuration, the full table with all configurations and standard deviation is shown and discussed in \cref{appendix:full_msl_smap_ablation_study}.
The row labeled "- contextual \dots" does not use the contextual hypersphere described in  \cref{subsec:contextual_hypersphere}, but instead a model trained using the original hypersphere classifier loss on the whole-window representation $\phi( \bm{w} ; \theta)$.
The contextual loss function provides a substantial performance boost, making our approach competitive
even without the data augmentation techniques. 
Each of the data augmentation techniques improves the performance further.
%
A further ablation study on the supervised Yahoo dataset can be found in \cref{table:benchmark_yahoo_supervised}.



\vspace{-5pt}
\subsection{Scaling from unsupervised to supervised}

To investigate how the performance of our approach changes as we scale from unsupervised, to semi-supervised, to fully supervised, we measure the performance of our approach as a function of the amount of ground truth labels on the \textsc{Yahoo} dataset, shown in \cref{fig:percentage_anomalie_v1}.
Firstly, we observe that the performance increases steadily with the amount of true anomaly labels, as desired. Secondly, by using synthetic anomalies (either \acrshort{po} or \acrshort{coe}), we can significantly boost the performance in the regime when no or only few labels are available. 
Finally, by using an injection technique that is well-aligned with the desired type of anomalies (\acrshort{po} in this case, as \textsc{Yahoo} contains a large number of single-point outliers), one can significantly improve performance over relying solely on the labeled data, this is explained by the very high class imbalance in anomaly detection. The flipside is, of course, that injecting anomalies that may be significantly different from the desired anomalies (\acrshort{coe} in this case) can ultimately hurt when enough labeled examples are available.

\subsection{Using specialized anomaly injection methods}
\label{subsec:specialized_anomalies}

\begin{figure*}[ht]
	\begin{subfigure}{.5\columnwidth}
		\caption{F1 score on the Performance first dimension of \acrshort{smap} with specialized anomaly injections.}
        \label{table:tunning_the_injected_anomalies}
        \centering
        \resizebox{0.96\columnwidth}{!}{%
        \begin{tabular}[h]{lccccccccccccc}
            \toprule
            Model & \acrshort{smap} 1st dimension \\
            \midrule
            \multicolumn{1}{l}{\acrshort{ncad}} 
            & 93.38 \\
            \multicolumn{1}{l}{\acrshort{ncad} + injections }
            & 96.48 \\
            \bottomrule
        \end{tabular}
        }
	\end{subfigure}
	\begin{subfigure}{.5\columnwidth}
    	\centering 
        \caption{F1 score vs.\ width of true anomalies for models trained only on point outliers, with different fractions of training examples mixed-up.
          }
        \includegraphics[width=.99\columnwidth]{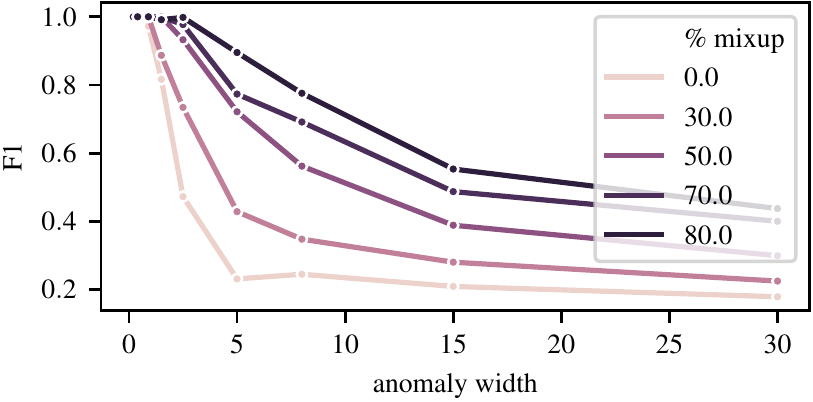}
        \label{fig:mixup_rate}
	\end{subfigure}
	\vspace*{-6mm}
	\caption{Investigating the potential of anomaly injections taking advantage of domain knowledge, and investigating the generalization of the model form \acrshort{po} when this knowledge is not available.}
\label{fig:synthetic_ts}
\end{figure*}

While in all our benchmarks we rely on completely generic anomalies for injection (\acrshort{coe} and \acrshort{po}), a by-product of our methodology is that the model can be guided towards detecting the desired class of anomalies by designing anomaly injection methods that mimic the true anomalies.
Designing such methods is often simple compared to finding enough examples of true anomalies as they are rare.
\Cref{table:tunning_the_injected_anomalies} demonstrates the effectiveness of this approach:
The first dimension of the \acrshort{smap} dataset contains 
slow slopes that are labeled as anomalous in the dataset. 
These are harder to detect for our model when only using \acrshort{coe} and \acrshort{po} \ because these cannot create similar behavior. 
We can design a simple anomaly injection that injects slopes to randomly selected region and labels it as anomalous. Training \acrshort{ncad} with these slopes gives a model that achieves a much better score.

This approach can be effective in applications where anomalies are subtle and closer to the normal data, and where some prior knowledge is available about the kind of anomalies that are to be detected.
However one may not have this prior knowledge or the resources required to create these injections.
This is a limitation of this technique which prevents it from being generally applicable. This is the reason why we did not use in for the comparison to the other methods.



\vspace{-5pt}
\subsection{Generalization from injected anomalies}

Artificial anomalies will always differ from the true anomalies to some extent, be it the ones created by \acrshort{coe}, \acrshort{po}, or more complex methods. 
This requires the model to bridge this gap and generalize from imperfect training examples to true anomalies.
By design the hypersphere formulation can help to bridge the gap, and we use \textsc{mixup} further improve the generalization capabilities of the model.
\Cref{fig:mixup_rate} shows the results of an experiment exploring one aspect of this generalization ability for \acrshort{ncad}. The model is trained with injected single-point outliers, and we measure the detection performance for anomalies of longer width.
For this experiment we use a synthetic base data set containing simple sinusoid time series with Gaussian noise.
We create multiple datasets from this base dataset adding true anomalies of varying width by convolving spike anomalies with Gaussian filters of different widths.
For training, regardless of the shape of the true anomalies, we use \acrshort{po} and train models using different \textsc{mixup}  rates, i.e., fraction of training examples with \textsc{mixup}  applied.
We observe that \textsc{mixup}  helps the model to generalize in this setting:
the higher the \textsc{mixup}  rate, the better the model generalizes to anomalies that differ from the injected examples, achieving higher F1 scores.
%


\vspace{-5pt}
\section{Discussion}
\label{sec:conclusions}

We present \acrshort{ncad}, a methodology for anomaly detection in time series that achieves state-of-the-art performance in a broad range of settings, including both the univariate and multivariate cases, as well as across the unsupervised, semi-supervised, and supervised anomaly detection regimes.
We demonstrate that combining expressive neural representation for time series with data augmentation techniques can outperform traditional approaches such as predictive models or methods based on reconstruction error.

%

We do not foresee clear potential negative societal impact of this work.
Time series anomaly detection is a general problem which is applied in many different domains, such as cyber-security where it can be used to automatically prevent attacks to power plants or hospitals.
While the anomaly detection results of our approach are good, we think that the detection of the algorithm should not be blindly followed in medical application impacting directly the patients health. 

\vspace{30pt}

\bibliography{references}

\begin{thebibliography}{45}
\providecommand{\natexlab}[1]{#1}
\providecommand{\url}[1]{\texttt{#1}}
\expandafter\ifx\csname urlstyle\endcsname\relax
  \providecommand{\doi}[1]{doi: #1}\else
  \providecommand{\doi}{doi: \begingroup \urlstyle{rm}\Url}\fi

\bibitem[kpi()]{kpi1}
Aiops challenge.
\newblock \url{http://iops.ai/dataset_detail/?id=10}.

\bibitem[Ayed et~al.(2020)Ayed, Stella, Januschowski, and
  Gasthaus]{ayed2020anomaly}
Ayed, F., Stella, L., Januschowski, T., and Gasthaus, J.
\newblock Anomaly detection at scale: The case for deep distributional time
  series models.
\newblock \emph{arXiv preprint arXiv:2007.15541}, 2020.

\bibitem[Bai et~al.(2018)Bai, Kolter, and Koltun]{Bai2018tcn}
Bai, S., Kolter, J.~Z., and Koltun, V.
\newblock {An Empirical Evaluation of Generic Convolutional and Recurrent
  Networks for Sequence Modeling}.
\newblock \emph{arXiv preprint arXiv:1803.01271}, mar 2018.
\newblock URL \url{http://arxiv.org/abs/1803.01271}.

\bibitem[Chandola et~al.(2009)Chandola, Banerjee, and Kumar]{anom4}
Chandola, V., Banerjee, A., and Kumar, V.
\newblock Anomaly detection: A survey.
\newblock \emph{ACM computing surveys (CSUR)}, 41\penalty0 (3):\penalty0 1--58,
  2009.

\bibitem[Deecke et~al.(2020)Deecke, Ruff, Vandermeulen, and
  Bilen]{Deecke2020residual}
Deecke, L., Ruff, L., Vandermeulen, R.~A., and Bilen, H.
\newblock {Deep Anomaly Detection by Residual Adaptation}.
\newblock \emph{arXiv preprint arXiv:2010.02310}, oct 2020.
\newblock URL \url{http://arxiv.org/abs/2010.02310}.

\bibitem[Falcon(2019)]{falcon2019pytorch}
Falcon, W.~A.
\newblock {PyTorch Lightning}, 2019.
\newblock URL \url{https://github.com/PyTorchLightning/pytorch-lightning}.

\bibitem[Franceschi et~al.(2019)Franceschi, Dieuleveut, and
  Jaggi]{Franceschi2019}
Franceschi, J.~Y., Dieuleveut, A., and Jaggi, M.
\newblock {Unsupervised scalable representation learning for multivariate time
  series}.
\newblock In \emph{Proceedings of the 33rd Conference on Neural Information
  Processing Systems, NeurIPS 2019}, volume~32, jan 2019.

\bibitem[Gao et~al.(2020)Gao, Song, Wen, Wang, Sun, and Xu]{Gao2020robustad}
Gao, J., Song, X., Wen, Q., Wang, P., Sun, L., and Xu, H.
\newblock {RobustTAD: Robust Time Series Anomaly Detection via Decomposition
  and Convolutional Neural Networks}.
\newblock \emph{arXiv preprint arXiv:2002.09545}, feb 2020.
\newblock URL \url{http://arxiv.org/abs/2002.09545}.

\bibitem[Guha et~al.(2016)Guha, Mishra, Roy, and Schrijvers]{rcf}
Guha, S., Mishra, N., Roy, G., and Schrijvers, O.
\newblock Robust random cut forest based anomaly detection on streams.
\newblock In \emph{International conference on machine learning}, pp.\
  2712--2721. PMLR, 2016.

\bibitem[Harris et~al.(2020)Harris, Millman, van~der Walt, Gommers, Virtanen,
  Cournapeau, Wieser, Taylor, Berg, Smith, Kern, Picus, Hoyer, van Kerkwijk,
  Brett, Haldane, del R{\'{i}}o, Wiebe, Peterson, G{\'{e}}rard-Marchant,
  Sheppard, Reddy, Weckesser, Abbasi, Gohlke, and Oliphant]{harris2020array}
Harris, C.~R., Millman, K.~J., van~der Walt, S.~J., Gommers, R., Virtanen, P.,
  Cournapeau, D., Wieser, E., Taylor, J., Berg, S., Smith, N.~J., Kern, R.,
  Picus, M., Hoyer, S., van Kerkwijk, M.~H., Brett, M., Haldane, A., del
  R{\'{i}}o, J.~F., Wiebe, M., Peterson, P., G{\'{e}}rard-Marchant, P.,
  Sheppard, K., Reddy, T., Weckesser, W., Abbasi, H., Gohlke, C., and Oliphant,
  T.~E.
\newblock Array programming with {NumPy}.
\newblock \emph{Nature}, 585\penalty0 (7825):\penalty0 357--362, September
  2020.
\newblock \doi{10.1038/s41586-020-2649-2}.
\newblock URL \url{https://doi.org/10.1038/s41586-020-2649-2}.

\bibitem[Hendrycks et~al.(2019)Hendrycks, Mazeika, and
  Dietterich]{Hendrycks2019oe}
Hendrycks, D., Mazeika, M., and Dietterich, T.
\newblock {Deep anomaly detection with outlier exposure}.
\newblock In \emph{Proceedings of the 7th International Conference on Learning
  Representations, ICLR 2019}, dec 2019.
\newblock URL \url{http://arxiv.org/abs/1812.04606}.

\bibitem[Hundman et~al.(2018)Hundman, Constantinou, Laporte, Colwell, and
  Soderstrom]{Hundman2018telemanon}
Hundman, K., Constantinou, V., Laporte, C., Colwell, I., and Soderstrom, T.
\newblock Detecting spacecraft anomalies using lstms and nonparametric dynamic
  thresholding.
\newblock In \emph{Proceedings of the 24th ACM SIGKDD international conference
  on knowledge discovery \& data mining}, pp.\  387--395, 2018.

\bibitem[Lavin \& Ahmad(2015)Lavin and Ahmad]{numenta}
Lavin, A. and Ahmad, S.
\newblock Evaluating real-time anomaly detection algorithms--the numenta
  anomaly benchmark.
\newblock In \emph{2015 IEEE 14th International Conference on Machine Learning
  and Applications (ICMLA)}, pp.\  38--44. IEEE, 2015.

\bibitem[Liberty et~al.(2020)Liberty, Karnin, Xiang, Rouesnel, Coskun,
  Nallapati, Delgado, Sadoughi, Astashonok, Das, Balioglu, Chakravarty, Jha,
  Gautier, Arpin, Januschowski, Flunkert, Wang, Gasthaus, Stella, Rangapuram,
  Salinas, Schelter, and Smola]{Liberty2020sagemaker}
Liberty, E., Karnin, Z., Xiang, B., Rouesnel, L., Coskun, B., Nallapati, R.,
  Delgado, J., Sadoughi, A., Astashonok, Y., Das, P., Balioglu, C.,
  Chakravarty, S., Jha, M., Gautier, P., Arpin, D., Januschowski, T., Flunkert,
  V., Wang, Y., Gasthaus, J., Stella, L., Rangapuram, S., Salinas, D.,
  Schelter, S., and Smola, A.
\newblock Elastic machine learning algorithms in amazon sagemaker.
\newblock In \emph{Proceedings of the 2020 ACM SIGMOD International Conference
  on Management of Data}, pp.\  731--737, 2020.

\bibitem[Lin et~al.(2020)Lin, Goyal, Girshick, He, and Dollar]{Lin2020focal}
Lin, T.-Y., Goyal, P., Girshick, R., He, K., and Dollar, P.
\newblock {Focal Loss for Dense Object Detection}.
\newblock \emph{IEEE Transactions on Pattern Analysis and Machine
  Intelligence}, 42\penalty0 (2):\penalty0 318--327, feb 2020.
\newblock ISSN 0162-8828.
\newblock \doi{10.1109/TPAMI.2018.2858826}.
\newblock URL \url{http://arxiv.org/abs/1708.02002
  https://ieeexplore.ieee.org/document/8417976/}.

\bibitem[Liu et~al.(2015)Liu, Zhao, Xu, Sun, Pei, Luo, Jing, and
  Feng]{liu2015opprentice}
Liu, D., Zhao, Y., Xu, H., Sun, Y., Pei, D., Luo, J., Jing, X., and Feng, M.
\newblock Opprentice: Towards practical and automatic anomaly detection through
  machine learning.
\newblock In \emph{Proceedings of the 2015 Internet Measurement Conference},
  pp.\  211--224, 2015.

\bibitem[Liznerski et~al.(2020)Liznerski, Ruff, Vandermeulen, Franks, Kloft,
  and M{\"{u}}ller]{Liznerski2020explainablead}
Liznerski, P., Ruff, L., Vandermeulen, R.~A., Franks, B.~J., Kloft, M., and
  M{\"{u}}ller, K.-R.
\newblock {Explainable Deep One-Class Classification}.
\newblock In \emph{Proceedings of the 9th International Conference on Learning
  Representations, ICLR 2021}, jul 2020.
\newblock URL \url{https://github.com/liznerski/fcdd
  http://arxiv.org/abs/2007.01760}.

\bibitem[Mathur \& Tippenhauer(2016)Mathur and Tippenhauer]{mathur2016swat}
Mathur, A.~P. and Tippenhauer, N.~O.
\newblock Swat: a water treatment testbed for research and training on ics
  security.
\newblock In \emph{2016 international workshop on cyber-physical systems for
  smart water networks (CySWater)}, pp.\  31--36. IEEE, 2016.

\bibitem[Pang et~al.(2020)Pang, Shen, Cao, and Hengel]{pang2020deep}
Pang, G., Shen, C., Cao, L., and Hengel, A. v.~d.
\newblock Deep learning for anomaly detection: A review.
\newblock \emph{arXiv preprint arXiv:2007.02500}, 2020.

\bibitem[Park et~al.(2018)Park, Hoshi, and Kemp]{lstmvae}
Park, D., Hoshi, Y., and Kemp, C.~C.
\newblock A multimodal anomaly detector for robot-assisted feeding using an
  lstm-based variational autoencoder.
\newblock \emph{IEEE Robotics and Automation Letters}, 3\penalty0 (3):\penalty0
  1544--1551, 2018.

\bibitem[Paszke et~al.(2019)Paszke, Gross, Massa, Lerer, Bradbury, Chanan,
  Killeen, Lin, Gimelshein, Antiga, Desmaison, K{\"{o}}pf, Yang, DeVito,
  Raison, Tejani, Chilamkurthy, Steiner, Fang, Bai, and
  Chintala]{Paszke2019pytorch}
Paszke, A., Gross, S., Massa, F., Lerer, A., Bradbury, J., Chanan, G., Killeen,
  T., Lin, Z., Gimelshein, N., Antiga, L., Desmaison, A., K{\"{o}}pf, A., Yang,
  E., DeVito, Z., Raison, M., Tejani, A., Chilamkurthy, S., Steiner, B., Fang,
  L., Bai, J., and Chintala, S.
\newblock {PyTorch: An Imperative Style, High-Performance Deep Learning
  Library}.
\newblock \emph{Proceedings of the 33rd Conference on Neural Information
  Processing Systems, NeurIPS 2019}, dec 2019.
\newblock URL \url{http://arxiv.org/abs/1912.01703}.

\bibitem[Perrone et~al.(2020)Perrone, Shen, Zolic, Shcherbatyi, Ahmed, Bansal,
  Donini, Winkelmolen, Jenatton, Faddoul, Pogorzelska, Miladinovic, Kenthapadi,
  Seeger, and Archambeau]{perrone2020smtuner}
Perrone, V., Shen, H., Zolic, A., Shcherbatyi, I., Ahmed, A., Bansal, T.,
  Donini, M., Winkelmolen, F., Jenatton, R., Faddoul, J.~B., Pogorzelska, B.,
  Miladinovic, M., Kenthapadi, K., Seeger, M., and Archambeau, C.
\newblock {Amazon SageMaker Automatic Model Tuning: Scalable Black-box
  Optimization}.
\newblock Technical report, Amazon, dec 2020.
\newblock URL \url{http://arxiv.org/abs/2012.08489}.

\bibitem[Ren et~al.(2019)Ren, Xu, Wang, Yi, Huang, Kou, Xing, Yang, Tong, and
  Zhang]{Ren2019srcnn}
Ren, H., Xu, B., Wang, Y., Yi, C., Huang, C., Kou, X.-A., Xing, T., Yang, M.,
  Tong, J., and Zhang, Q.
\newblock {Time-Series Anomaly Detection Service at Microsoft}.
\newblock In \emph{Proceedings of the 25th ACM SIGKDD International Conference
  on Knowledge Discovery {\&} Data Mining}, volume~19, pp.\  3009--3017, New
  York, NY, USA, jul 2019. ACM.
\newblock ISBN 9781450362016.
\newblock \doi{10.1145/3292500.3330680}.
\newblock URL \url{https://doi.org/10.1145/3292500.3330680
  https://dl.acm.org/doi/10.1145/3292500.3330680}.

\bibitem[Ruff et~al.(2018)Ruff, Vandermeulen, G{\"{o}}rnitz, Deecke, Siddiqui,
  Binder, Uller, and Kloft]{Ruff2018deepsvd}
Ruff, L., Vandermeulen, R.~A., G{\"{o}}rnitz, N., Deecke, L., Siddiqui, S.~A.,
  Binder, A., Uller, E.~M., and Kloft, M.
\newblock {Deep One-Class Classification}.
\newblock In \emph{Proceedings of the 35th International Conference on Machine
  Learning, ICML 2018}, pp.\  4393----4402, 2018.
\newblock URL \url{http://proceedings.mlr.press/v80/ruff18a}.

\bibitem[Ruff et~al.(2020{\natexlab{a}})Ruff, Vandermeulen, Franks,
  M{\"{u}}ller, and Kloft]{Ruff2020hsc}
Ruff, L., Vandermeulen, R.~A., Franks, B.~J., M{\"{u}}ller, K.-R., and Kloft,
  M.
\newblock {Rethinking Assumptions in Deep Anomaly Detection}.
\newblock \emph{arXiv preprint arXiv:2006.00339}, may 2020{\natexlab{a}}.
\newblock URL \url{http://arxiv.org/abs/2006.00339}.

\bibitem[Ruff et~al.(2020{\natexlab{b}})Ruff, Vandermeulen, G{\"{o}}rnitz,
  Binder, M{\"{u}}ller, M{\"{u}}ller, and Kloft]{Ruff2020deepsad}
Ruff, L., Vandermeulen, R.~A., G{\"{o}}rnitz, N., Binder, A., M{\"{u}}ller, E.,
  M{\"{u}}ller, K.-R., and Kloft, M.
\newblock {Deep Semi-Supervised Anomaly Detection}.
\newblock In \emph{Proceedings of the 8th International Conference on Learning
  Representations, ICLR 2020}, jun 2020{\natexlab{b}}.
\newblock URL \url{http://arxiv.org/abs/1906.02694}.

\bibitem[Ruff et~al.(2021)Ruff, Kauffmann, Vandermeulen, Montavon, Samek,
  Kloft, Dietterich, and Muller]{Ruff2020adreview}
Ruff, L., Kauffmann, J.~R., Vandermeulen, R.~A., Montavon, G., Samek, W.,
  Kloft, M., Dietterich, T.~G., and Muller, K.-R.
\newblock {A Unifying Review of Deep and Shallow Anomaly Detection}.
\newblock \emph{Proceedings of the IEEE}, 109\penalty0 (5):\penalty0 756--795,
  may 2021.
\newblock ISSN 0018-9219.
\newblock \doi{10.1109/JPROC.2021.3052449}.
\newblock URL \url{https://www.statista.com/ http://arxiv.org/abs/2009.11732
  https://ieeexplore.ieee.org/document/9347460/}.

\bibitem[Schlegl et~al.(2017)Schlegl, Seeb{\"o}ck, Waldstein, Schmidt-Erfurth,
  and Langs]{anogan}
Schlegl, T., Seeb{\"o}ck, P., Waldstein, S.~M., Schmidt-Erfurth, U., and Langs,
  G.
\newblock Unsupervised anomaly detection with generative adversarial networks
  to guide marker discovery.
\newblock In \emph{International conference on information processing in
  medical imaging}, pp.\  146--157. Springer, 2017.

\bibitem[Shen et~al.(2020)Shen, Li, and Kwok]{thoc}
Shen, L., Li, Z., and Kwok, J.
\newblock Timeseries anomaly detection using temporal hierarchical one-class
  network.
\newblock \emph{Advances in Neural Information Processing Systems}, 33, 2020.

\bibitem[Shewhart(1931)]{shewhart1931economic}
Shewhart, W.~A.
\newblock \emph{Economic control of quality of manufactured product}.
\newblock Macmillan And Co Ltd, London, 1931.

\bibitem[Shipmon et~al.(2017)Shipmon, Gurevitch, Piselli, and
  Edwards]{shipmon2017time}
Shipmon, D.~T., Gurevitch, J.~M., Piselli, P.~M., and Edwards, S.~T.
\newblock Time series anomaly detection; detection of anomalous drops with
  limited features and sparse examples in noisy highly periodic data.
\newblock \emph{arXiv preprint arXiv:1708.03665}, 2017.

\bibitem[Siffer et~al.(2017)Siffer, Fouque, Termier, and Largouet]{dspot}
Siffer, A., Fouque, P.-A., Termier, A., and Largouet, C.
\newblock Anomaly detection in streams with extreme value theory.
\newblock In \emph{Proceedings of the 23rd ACM SIGKDD International Conference
  on Knowledge Discovery and Data Mining}, pp.\  1067--1075, 2017.

\bibitem[Smolyakov et~al.(2019)Smolyakov, Sviridenko, Ishimtsev, Burikov, and
  Burnaev]{smolyakovLearningEnsemblesAnomaly2019}
Smolyakov, D., Sviridenko, N., Ishimtsev, V., Burikov, E., and Burnaev, E.
\newblock Learning {{Ensembles}} of {{Anomaly Detectors}} on {{Synthetic
  Data}}.
\newblock \emph{arXiv:1905.07892 [cs, stat]}, May 2019.

\bibitem[Su et~al.(2019)Su, Zhao, Niu, Liu, Sun, and Pei]{omnianomaly}
Su, Y., Zhao, Y., Niu, C., Liu, R., Sun, W., and Pei, D.
\newblock Robust anomaly detection for multivariate time series through
  stochastic recurrent neural network.
\newblock In \emph{Proceedings of the 25th ACM SIGKDD International Conference
  on Knowledge Discovery \& Data Mining}, pp.\  2828--2837, 2019.

\bibitem[Tatbul et~al.(2018)Tatbul, Lee, Zdonik, Alam, and
  Gottschlich]{tatbul2018precision}
Tatbul, N., Lee, T.~J., Zdonik, S., Alam, M., and Gottschlich, J.
\newblock Precision and recall for time series.
\newblock \emph{Advances in Neural Information Processing Systems},
  31:\penalty0 1920--1930, 2018.

\bibitem[Tax \& Duin(2004)Tax and Duin]{svdd}
Tax, D.~M. and Duin, R.~P.
\newblock Support vector data description.
\newblock \emph{Machine learning}, 54\penalty0 (1):\penalty0 45--66, 2004.

\bibitem[Tuluptceva et~al.(2020)Tuluptceva, Bakker, Fedulova, Schulz, and
  Dylov]{Tuluptceva2020ad}
Tuluptceva, N., Bakker, B., Fedulova, I., Schulz, H., and Dylov, D.~V.
\newblock {Anomaly Detection with Deep Perceptual Autoencoders}.
\newblock \emph{arXiv preprint arXiv:2006.13265}, jun 2020.
\newblock URL \url{http://arxiv.org/abs/2006.13265}.

\bibitem[van~den Oord et~al.(2016)van~den Oord, Dieleman, Zen, Simonyan,
  Vinyals, Graves, Kalchbrenner, Senior, and Kavukcuoglu]{Oord2016wavenet}
van~den Oord, A., Dieleman, S., Zen, H., Simonyan, K., Vinyals, O., Graves, A.,
  Kalchbrenner, N., Senior, A., and Kavukcuoglu, K.
\newblock {WaveNet: A Generative Model for Raw Audio}.
\newblock Technical report, Google, sep 2016.
\newblock URL \url{http://arxiv.org/abs/1609.03499
  https://research.google/pubs/pub45774/}.

\bibitem[Wu \& Keogh(2020)Wu and Keogh]{wu2020current}
Wu, R. and Keogh, E.~J.
\newblock Current time series anomaly detection benchmarks are flawed and are
  creating the illusion of progress.
\newblock \emph{arXiv preprint arXiv:2009.13807}, 2020.

\bibitem[Xu et~al.(2018)Xu, Chen, Zhao, Li, Bu, Li, Liu, Zhao, Pei, Feng,
  et~al.]{donut}
Xu, H., Chen, W., Zhao, N., Li, Z., Bu, J., Li, Z., Liu, Y., Zhao, Y., Pei, D.,
  Feng, Y., et~al.
\newblock Unsupervised anomaly detection via variational auto-encoder for
  seasonal kpis in web applications.
\newblock In \emph{Proceedings of the 2018 World Wide Web Conference}, pp.\
  187--196, 2018.

\bibitem[Zaheer et~al.(2018)Zaheer, Reddi, Sachan, Kale, and
  Kumar]{Zaheer2018yogi}
Zaheer, M., Reddi, S.~J., Sachan, D., Kale, S., and Kumar, S.
\newblock {Adaptive methods for nonconvex optimization}.
\newblock In \emph{Proceedings of the 32nd Conference on Neural Information
  Processing Systems, NIPS 2018}, volume 2018-Decem, pp.\  9793--9803, 2018.

\bibitem[Zhang et~al.(2019)Zhang, Song, Chen, Feng, Lumezanu, Cheng, Ni, Zong,
  Chen, and Chawla]{Zhang2019mscred}
Zhang, C., Song, D., Chen, Y., Feng, X., Lumezanu, C., Cheng, W., Ni, J., Zong,
  B., Chen, H., and Chawla, N.~V.
\newblock {A Deep Neural Network for Unsupervised Anomaly Detection and
  Diagnosis in Multivariate Time Series Data}.
\newblock \emph{Proceedings of the AAAI Conference on Artificial Intelligence},
  33:\penalty0 1409--1416, jul 2019.
\newblock ISSN 2374-3468.
\newblock \doi{10.1609/aaai.v33i01.33011409}.
\newblock URL \url{www.aaai.org
  https://aaai.org/ojs/index.php/AAAI/article/view/3942}.

\bibitem[Zhang et~al.(2018)Zhang, Cisse, Dauphin, and
  Lopez-Paz]{Zhang2018mixup}
Zhang, H., Cisse, M., Dauphin, Y.~N., and Lopez-Paz, D.
\newblock {Mixup: Beyond Empirical Risk Minimization}.
\newblock In \emph{Proceedings of the Sixth International Conference on
  Learning Representations, ICLR 2018}, oct 2018.

\bibitem[Zhao et~al.(2020)Zhao, Wang, Duan, Huang, Cao, Tong, Xu, Bai, Tong,
  and Zhang]{zhao2020multivariate}
Zhao, H., Wang, Y., Duan, J., Huang, C., Cao, D., Tong, Y., Xu, B., Bai, J.,
  Tong, J., and Zhang, Q.
\newblock Multivariate time-series anomaly detection via graph attention
  network.
\newblock \emph{arXiv preprint arXiv:2009.02040}, 2020.

\bibitem[Zong et~al.(2018)Zong, Song, Min, Cheng, Lumezanu, Cho, and
  Chen]{dagmm}
Zong, B., Song, Q., Min, M.~R., Cheng, W., Lumezanu, C., Cho, D., and Chen, H.
\newblock Deep autoencoding gaussian mixture model for unsupervised anomaly
  detection.
\newblock In \emph{International Conference on Learning Representations}, 2018.

\end{thebibliography}
\bibliographystyle{icml2021}

\clearpage
\newpage

\appendix

\section{Model Architecture}
\label{sec:ncad_architect}

\subsection{Encoder}
\label{sec:ncad_architect_encoder}
Our encoder function $g(\cdot): \mathbb{R}^{D \times L} \rightarrow \mathbb{R}^{E}$
\footnote{$D$ denotes the dimension of the time series ($D=1$ for univariate series), $L$ is the length
of the windows, and $E$ the dimension of the vector representation.}
is similar to the encoder proposed by \citet{Franceschi2019} for generating universal representations
of multivariate time series.

\begin{figure}[tbh]
    \centering
    \def\svgwidth{0.7\textwidth}
    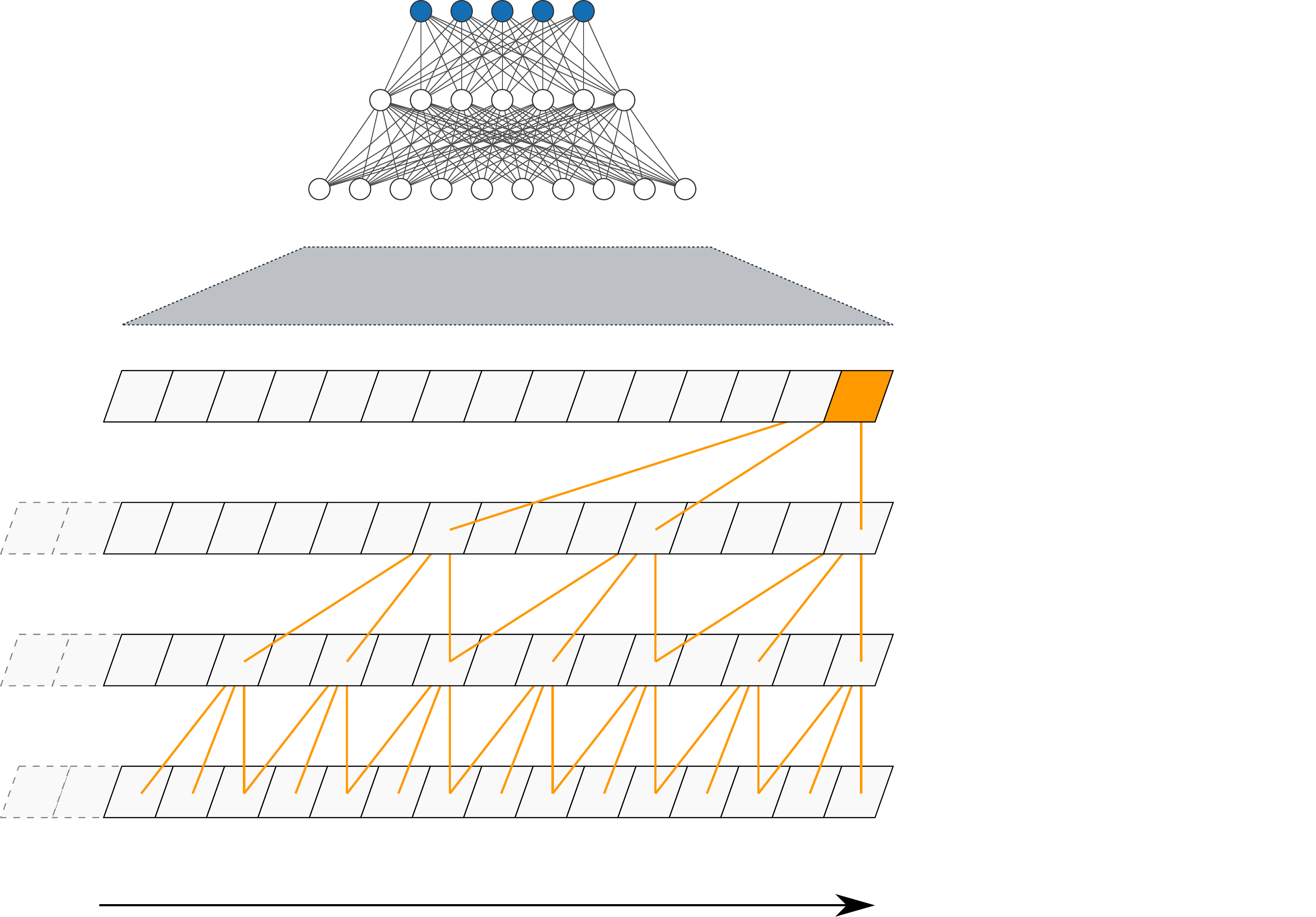
    \caption[]{Illustration of the encoder architecture.}
    \label{fig:ncad_encoder}
\end{figure}

The architecture is based on multi-stack \acrfullpl{tcn} \citep{Bai2018tcn},
which combines causal convolutions with residual connections. The output of this causal network is passed
to an adaptive max pooling layer, aggregating the temporal dimension into a fixed-size vector.\footnote{In our experiments, using adaptive pooling consistently outperformed the global pooling alternative.}
A linear transformation is applied to produce the unnormalized vector representations,
which are then $L2$-normalized to produce the final output of the encoder. See \cref{fig:ncad_encoder} for an illustration.

\subsection{Distance}
\label{sec:ncad_architect_dist}
We introduce a ``distance'' \footnote{The function $\dist(\cdot,\cdot)$ in our framework is not strictly a distance in the mathematical sense. Specifically, the triangle inequality is not required, but it is expected to be symmetric and non-negative.} function, $\dist(\cdot,\cdot) : \mathbb{R}^{E} \times \mathbb{R}^{E} \rightarrow \mathbb{R}^+$, as a central element in our approach. For a given window $x=(x_{(c)},x_{(s)})$, the distance function is used
to compare the embeddings of the entire window, $z=\phi(x;\theta)$, with the embedding of its corresponding context segment, $z_{(c)}=\phi(x_{(c)};\theta)$.
The output of this function can be directly interpreted as an anomaly score for the associated suspect segment $x_{(s)}$.

We explored two types of distances: the \emph{Euclidean distance},
\[
    \dist_{L2}(x,y) = \norm{x-y}_2 = \sqrt{ (x-y) \cdot (x-y) },
\]
and the \emph{Cosine distance}, defined as a simple logarithmic mapping of the cosine similarity to the positive reals,
\[
    \dist_{\cos}(x,y) = -\log\left(\frac{1+\cossim(x,y)}{2}\right)
\]
where $\cossim(x,y) = \dfrac{x \cdot y}{\Vert x \Vert _2 \cdot \Vert y \Vert _2}$ is the cosine similarity between $x$ and $y$.

In our experiments, we found that both distances were able to achieve state-of-the-art results in most of the reported benchmarks datasets, with slightly better performance from the Euclidean distance. All the results of \acrshort{ncad} reported in \cref{sec:experiments,appendix:experiments_extra} are based on the Euclidean distance. 

Moreover, the \acrshort{ncad} framework can be extended to use other distances, e.g. other $L_p$ norms, the pseudo-Hubber norm used in the \citet{Ruff2020hsc}, or even trainable neural-based distances.

\subsection{Probabilistic scoring function and Classification Loss}
\label{sec:ncad_architect_loss}

The final element in our model is the binary classification loss which measures the agreement between target labels $y$ and the assigned anomaly scores.

As described in \cref{sec:ncad}, for a given window $x=(x_{(c)},x_{(s)})$ and encoder $\phi(\cdot;\theta)$, we compute the anomaly score for the suspect window $x_{(s)}$, as the distance between the corresponding representations of the full window and the context window: $dist( \phi(x;\theta), \phi(x_{(c)};\theta))$.

This score is mapped into a pseudo-probability of an anomaly ($y=1$) via the probabilistic scoring function $\ell(\cdot)$,
\[
p = 1-\ell( dist( g(x), g(x_{(c)})) ),
\]
which is used within the \acrlong{bce} loss to define the target to minimize during training. We train the encoder $\phi(\cdot;\theta)$ using mini-batch gradient descent (see \cref{appendix:ncad_implementation}), taking $B$ randomly selected windows, and minimizing
\[
L_{BCE} = \frac{1}{B}\sum_{i=1}^{B} -\left[ y_i \; \log p_i + (1 - y_i) \; \log (1 - p_i) \right] \;.
\]

Alternative classification losses can be considered as an extension of the standard \acrshort{ncad} framework. For example, the \acrfull{mae},
the \acrfull{mse}, or the Focal Loss \citep{Lin2020focal}. These losses may be particularly useful in applications with significant contamination of labels.

\section{Data Augmentation}

As presented in \cref{sec:data_augmentation}, our framework relies on three data augmentation
methods for time series, we provide more details and some visualizations in this section.

\begin{figure*}[ht]
	\begin{subfigure}{.5\columnwidth}
    	\centering 
        \includegraphics[width=.99\columnwidth]{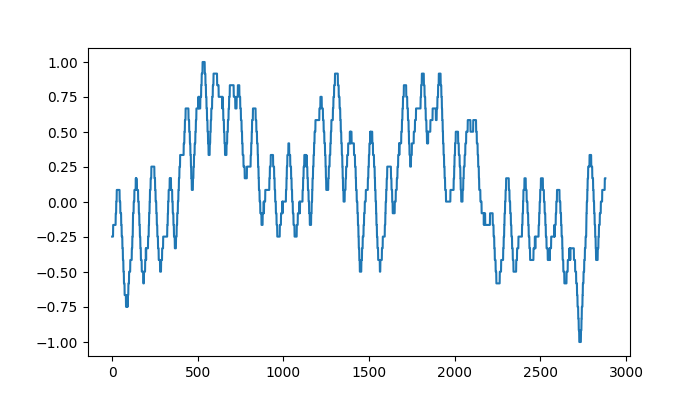}
	\end{subfigure}
	\begin{subfigure}{.5\columnwidth}
    	\centering 
        \includegraphics[width=.99\columnwidth]{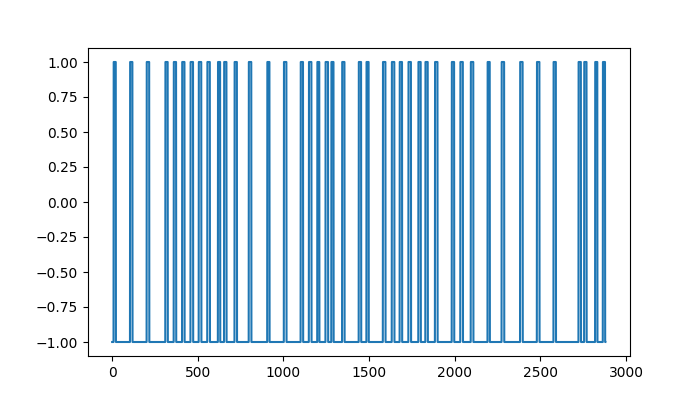}
	\end{subfigure}
	\begin{subfigure}{.5\columnwidth}
    	\centering 
        \includegraphics[width=.99\columnwidth]{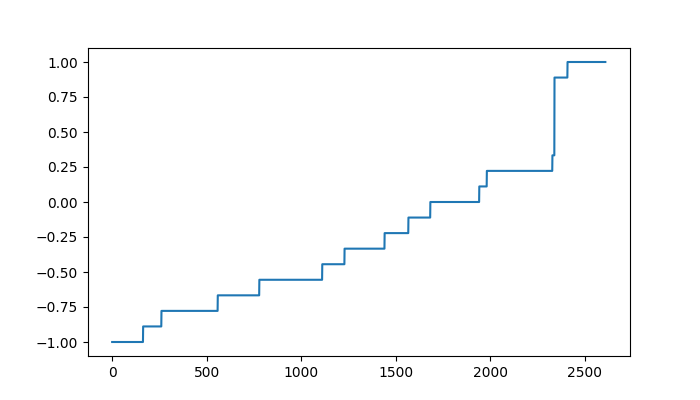}
	\end{subfigure}
	\begin{subfigure}{.5\columnwidth}
    	\centering 
        \includegraphics[width=.99\columnwidth]{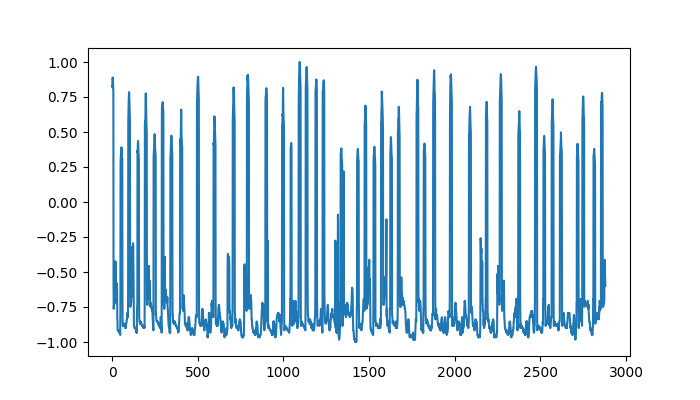}
	\end{subfigure}
\caption{The four original time series used for visualization of the data augmentation methods.}
\label{fig:all_the_empty_ts}
\end{figure*}

\Cref{fig:all_the_empty_ts} shows four time series drawn from the SMAP benchmark dataset, we use these to visualize each of the data augmentation method.

\subsection{Contextual Outlier Exposure (COE)}
\label{appendix:coe}

As described in \cref{sec:data_augmentation}, we use \acrshort{coe} at training time to generate additional anomalous training examples. These are created by replacing a chunk of values in one suspect window with values taken
from another suspect window in the same batch.

\begin{figure*}[ht]
	\begin{subfigure}{.5\columnwidth}
    	\centering 
        \includegraphics[width=.99\columnwidth]{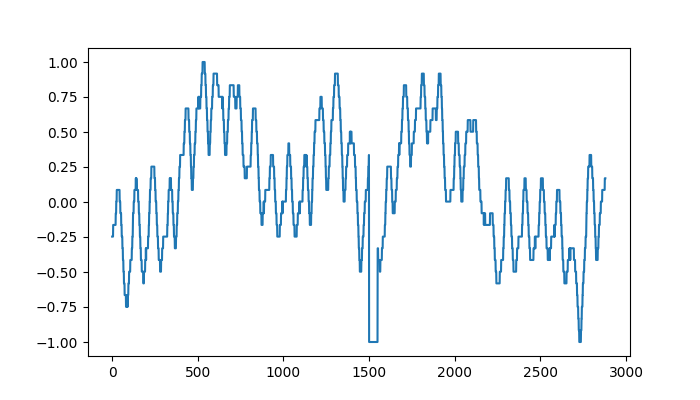}
        \caption{}
	\end{subfigure}
	\begin{subfigure}{.5\columnwidth}
    	\centering 
        \includegraphics[width=.99\columnwidth]{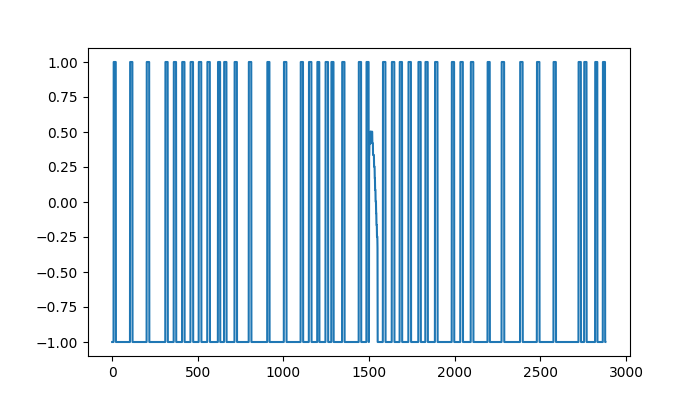}
        \caption{}
	\end{subfigure}
	\begin{subfigure}{.5\columnwidth}
    	\centering 
        \includegraphics[width=.99\columnwidth]{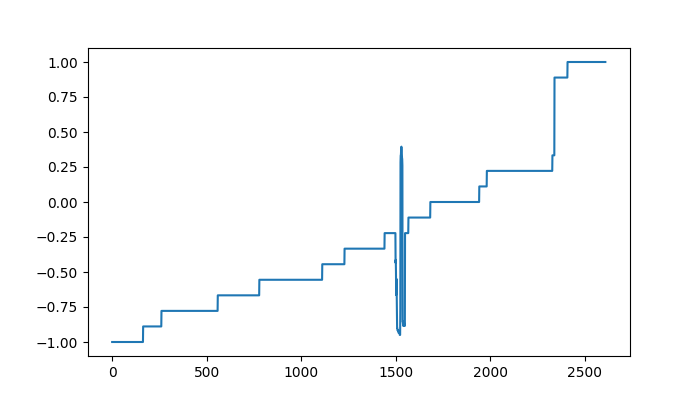}
        \caption{}
	\end{subfigure}
	\begin{subfigure}{.5\columnwidth}
    	\centering 
        \includegraphics[width=.99\columnwidth]{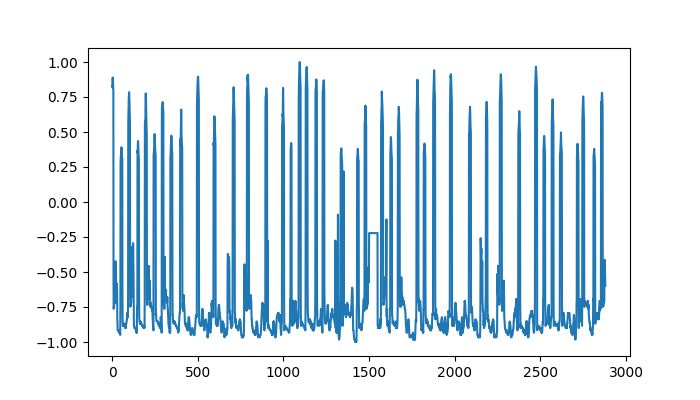}
        \caption{}
	\end{subfigure}
	\caption{Visualization of coe. Each of the series has its window between 1500 and 1550 swapped with its horizontal neighbor time series: (a) swaps with (b) and (c) swaps with (b)). }
\label{fig:all_the_coe_ts}
\end{figure*}

As an example, consider \cref{fig:all_the_coe_ts}, each time series has the window between 1500 and 1550 swapped with its horizontal neighbor. We can see that this creates an anomalous window that does not follow the expected behavior. In some cases, the swapping of values create jumps, as in (c); in other cases the change is more subtle, like in (d), where the series becomes constant for the duration of the interval or (a) and (b) where the regular pattern is broken.
%
\footnote{Our implementation of \acrshort{coe} can be found in file \path{src/ncad/model/outlier_exposure.py} of the supplementary code.}

\subsection{Point Outliers (PO)}
\label{appendix:po}

As described in \cref{sec:data_augmentation}, \acrlong{po} allow to add single isolated outliers to the time series. 
We simply add a spike to the time series at a random location in time. 
By default, the spike is between 0.5 and 3 time the inter-quartile range of the 100 points around the spike location.

\begin{figure*}[ht]
	\begin{subfigure}{.5\columnwidth}
    	\centering 
        \includegraphics[width=.99\columnwidth]{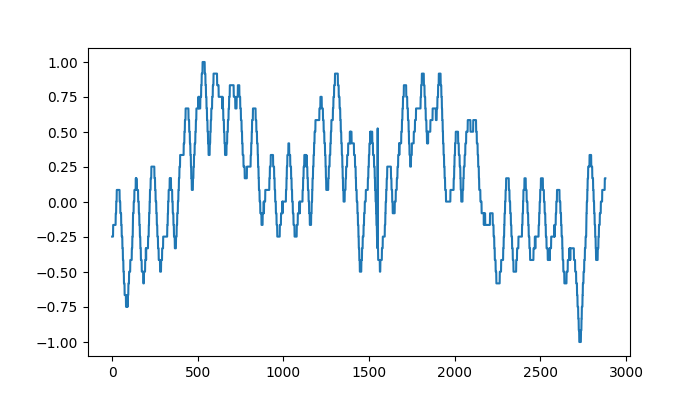}
	\end{subfigure}
	\begin{subfigure}{.5\columnwidth}
    	\centering 
        \includegraphics[width=.99\columnwidth]{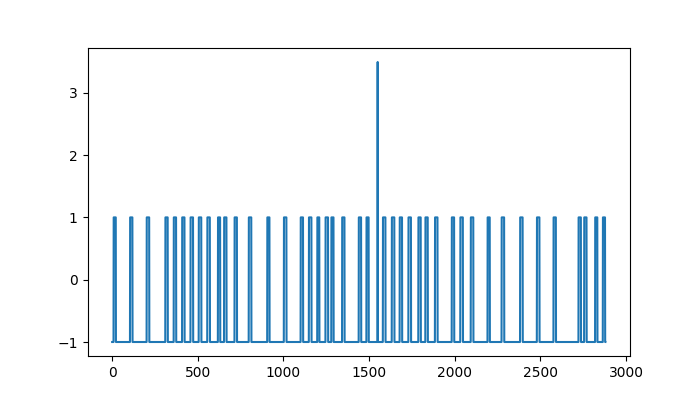}
	\end{subfigure}
	\begin{subfigure}{.5\columnwidth}
    	\centering 
        \includegraphics[width=.99\columnwidth]{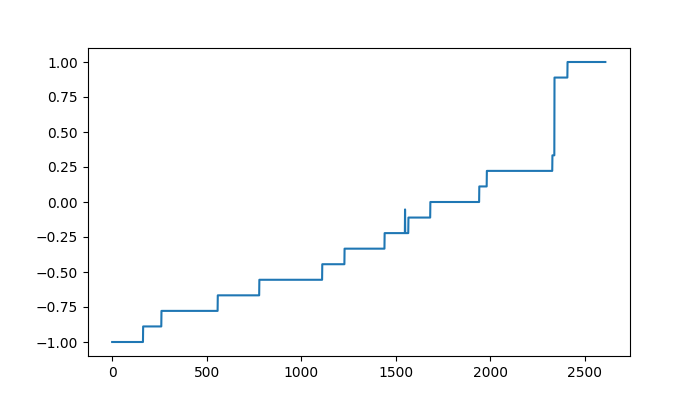}
	\end{subfigure}
	\begin{subfigure}{.5\columnwidth}
    	\centering 
        \includegraphics[width=.99\columnwidth]{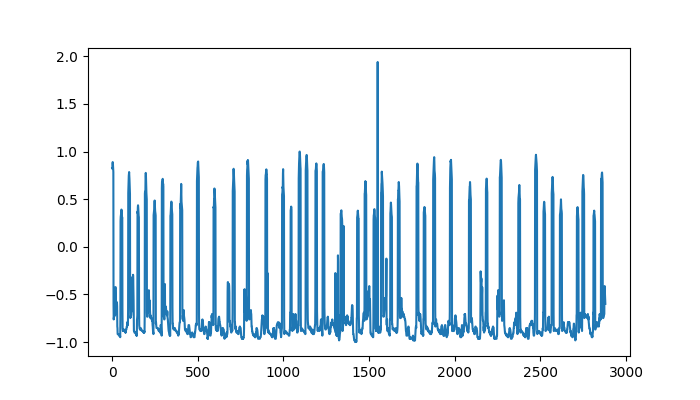}
	\end{subfigure}
	\caption{Visualization of po. In each of the time series a point outlier is injected at the 1550th value.}
    \label{fig:all_the_po_ts}
\end{figure*}

With this method, the injected spike can be a local outlier, but is not necessarily a global outlier as its magnitude could be within the range of other values in the time series.
Similarly to \acrshort{coe}, in the case of multivariate time series we select a random subset of dimensions on which we add the spike.
In \cref{fig:all_the_po_ts} we visualizes some examples of the injected point outliers. These are added to the 1550th value of each of the time series. We can see that they break the normal patterns but do not necessarily result in extreme events.
\footnote{Our implementation of \acrshort{po} can be found in file \path{src/ncad/ts/transforms/spikes_injection.py} of the supplementary code.}

\subsection{Time series Mixup}
\label{appendix:mixup}
As described in \cref{sec:data_augmentation}, inspired by \citet{Zhang2018mixup} we use an adapted version of the \textsc{Mixup} procedure as part of our framework.

We sample $\lambda \sim \text{Beta}(\alpha, \alpha)$ \footnote{we set $\alpha = 0.05$, as this value gave the best generalization among the values that were tried in the experience of \cref{fig:mixup_rate}}.
Using this $\lambda$ we create a new training example as a convex combination of two examples from the batch:
\begin{align*}
x_{\text{new}} &= \lambda x^{(i)} + (1 - \lambda) x^{(j)}, \hspace{40pt} \text{where $x^{(i)} $ and $x^{(j)}$ are two whole windows sampled from the batch}
\\
y_{\text{new}} &= \lambda y_s^{(i)}  + (1 - \lambda) y_s^{(j)}, \hspace{40pt} \text{where $y_s^{(i)} $ and $y_s^{(j)}$ are the two corresponding labels. }
\end{align*}
Note that, in addition to the new time series values $x_{\text{new}}$, the method also produces soft labels $y_{\text{new}}$, different to 0 or 1, which are used during training.
\footnote{Our implementation of mixup can be found in file \path{src/ncad/model/mixup.py} of the supplementary code.}

Figure \ref{fig:all_the_mixup_ts} shows example time series created using mixup. Each of the original time series is mixed up with its horizontal neighbor time series. We see that the newly created series have characteristics from both time series to create a new realistic time series. The patterns in (a) and (b) became a bit more noisy. The slope of (c) has the additional spiky pattern from (d) and the pattern in (d) now slowly ramps up.

\begin{figure*}[ht]
	\begin{subfigure}{.5\columnwidth}
    	\centering 
        \includegraphics[width=.99\columnwidth]{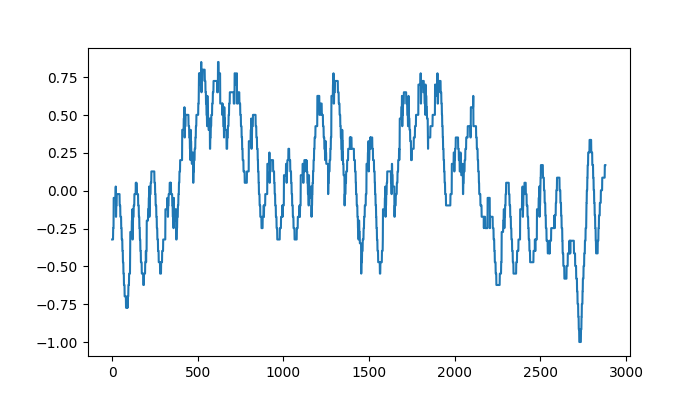}
        \caption{}
	\end{subfigure}
	\begin{subfigure}{.5\columnwidth}
    	\centering 
        \includegraphics[width=.99\columnwidth]{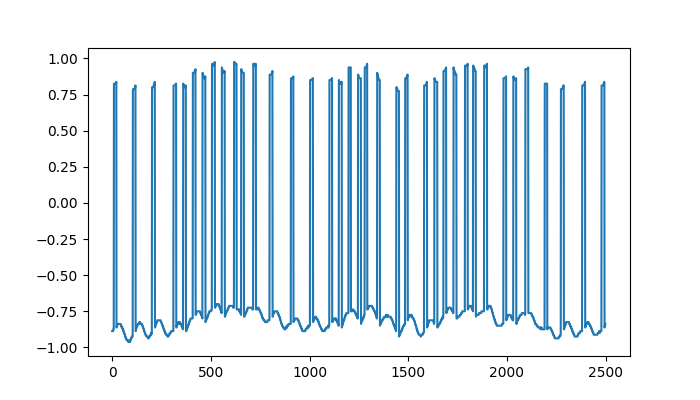}
        \caption{}
	\end{subfigure}
	\begin{subfigure}{.5\columnwidth}
    	\centering 
        \includegraphics[width=.99\columnwidth]{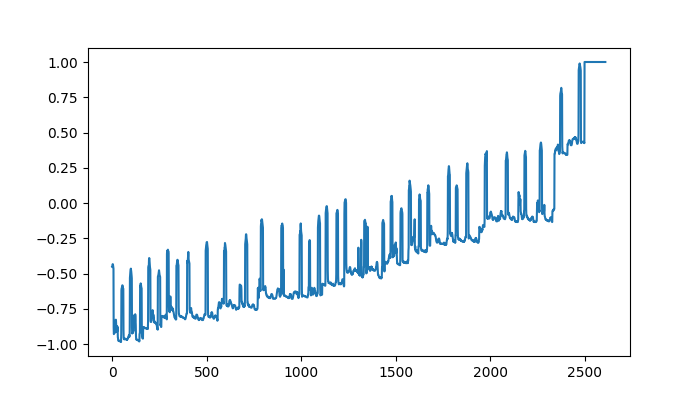}
        \caption{}
	\end{subfigure}
	\begin{subfigure}{.5\columnwidth}
    	\centering 
        \includegraphics[width=.99\columnwidth]{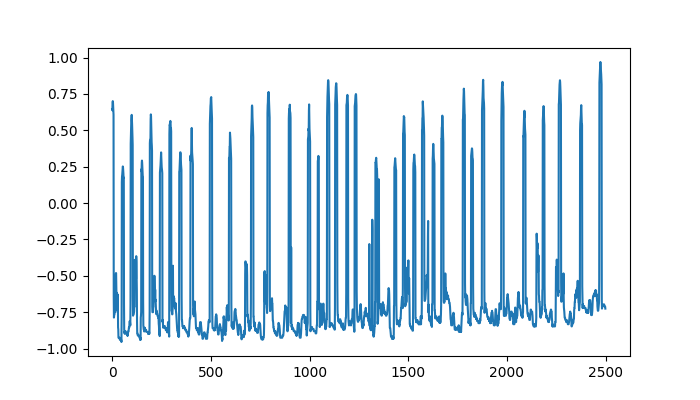}
        \caption{}
	\end{subfigure}
	\caption{Visualization of time series mixup. Each of the series is "mixed up" with the its horizontal neighbor time series: (a) with (b) and (c) with (d)). }
    \label{fig:all_the_mixup_ts}
\end{figure*}

\section{Further Results and Ablation Studies}
\label{appendix:experiments_extra}

\subsection{Ablation Study on SMAP and MSL}
\label{appendix:full_msl_smap_ablation_study}

Here we present a full ablation study on the \acrshort{smap} and \acrshort{msl} datasets. We consider variations of the framework by removing some of its components, and train the model in each configurations twice. We report the average and standard deviation of these runs.

First, we observed that the contextual hypersphere formulation improves performance of the model. In the setting with all the data augmentation techniques "- contextual" the difference is not very big 1.98\% F1 and 1.17\% F1 on SMAP and MSL respectively. However, in the setting where none of the data augmentation is used, it makes a dramatic difference to use this formulation: 11.81\% F1 and 43.44\% F1 on SMAP and MSL respectively.
Further, we can see that solely with the contextual hypersphere and without relying on any data augmentation technique the model can achieve a very reasonable performance.

\begin{table}[H]
  \caption{F1 score of the model on \acrshort{smap} and \acrshort{msl}}
  \label{table:ablation}
  \centering
  \begin{tabular}{lccccccccccccc}
    \toprule
        Model & \acrshort{smap} & MSL \\
    \midrule
        THOC \citep{thoc}&  95.18 & 93.67 \\
    \midrule
        \multicolumn{1}{l}{\acrshort{ncad} w/ \acrshort{coe}, \acrshort{po}, mixup}  & 94.45 $\pm$ 0.68 &  95.60 $\pm$ 0.68  \\
        \multicolumn{1}{l}{\hspace{6pt} - \acrshort{coe}}  & 88.59  $ \pm $  1.81  & 94.66  $ \pm $  0.22 \\
        \multicolumn{1}{l}{\hspace{6pt} - \acrshort{po} }  & 94.28  $ \pm $  0.45  & 94.73  $ \pm $  0.35 \\
        \multicolumn{1}{l}{\hspace{6pt} - mixup}  & 92.69  $ \pm $  1.14 &  95.59  $ \pm $  0.01  \\
        \multicolumn{1}{l}{\hspace{6pt} - mixup - \acrshort{po}}  &  94.4  $ \pm $  0.43 & 94.12  $ \pm $  0.77  \\
        \multicolumn{1}{l}{\hspace{6pt} - mixup - \acrshort{coe}}  & 86.86  $ \pm $  0.7 &  91.7  $ \pm $  2.58 \\
        \multicolumn{1}{l}{\hspace{6pt}  - \acrshort{coe} - \acrshort{po} }  &  60.48  $ \pm $  9.7  & 42.02 $ \pm $ 6.34 \\
        \multicolumn{1}{l}{\hspace{6pt} - mixup - \acrshort{coe} - \acrshort{po} }  &  66.9  $ \pm $  2.01 & 79.47  $ \pm $  9.39 \\
        \multicolumn{1}{l}{\hspace{6pt} - contextual}  &  92.47  $ \pm $  0.53  & 94.43  $ \pm $  0.15  \\
        \multicolumn{1}{l}{\hspace{6pt} - contextual - \acrshort{coe}}  &  91.86  $ \pm $  0.96 & 88.29  $ \pm $  0.43 \\
        \multicolumn{1}{l}{\hspace{6pt} - contextual - \acrshort{po} }  & 93.39  $ \pm $ 0.61 & 90.68  $ \pm $  0.74 \\
        \multicolumn{1}{l}{\hspace{6pt} - contextual - mixup}  & 94.37  $ \pm $  0.21 & 95.07  $ \pm $  0.14 \\
        \multicolumn{1}{l}{\hspace{6pt} - contextual - mixup - \acrshort{po}}  & 93.24  $ \pm $  0.31  & 90.89  $ \pm $  0.46   \\
        \multicolumn{1}{l}{\hspace{6pt} - contextual - mixup - \acrshort{coe}}  & 89.88  $ \pm $  2.53 & 87.26  $ \pm $  4.17 \\
        \multicolumn{1}{l}{\hspace{6pt} - contextual - \acrshort{coe} - \acrshort{po} }  & 54.95  $ \pm $  2.62 & 32.05  $ \pm $  0.17  \\
        \multicolumn{1}{l}{\hspace{6pt} - contextual - \acrshort{coe} - mixup - \acrshort{po} }  & 55.09  $ \pm $  1.0 & 36.03  $ \pm $  3.01 \\
    \bottomrule
    \end{tabular}
\end{table}

We also observed that both \acrshort{coe} and \acrshort{po} jointly improve the model performance.
If we remove separately one of these elements, neither of the two tends to have a large impact on the performance. However, if none of them is used the performance drops drastically: by 43.98\% F1 and 53.58\% F1 for SMAP and MSL respectively, and the drop is even bigger when not using the contextual inductive bias.

It is interesting to note that it does not seem to be a good idea to use mixup as the only data augmentation method (at least in this unsupervised setting).
In the setting where neither \acrshort{coe} nor \acrshort{po} are used, using mixup seems to significantly deteriorate the performance: by 6.42\% and 37.45\% on SMAP and MSL respectively.
We conjecture that this is due to the fact that, for this datasets, there are no labels in the training data and so mixup does not allow to create soft-labels. In addition, mixup creates new time series that may not correspond to the original data distribution, these may deviate the learning away from the original data distribution.

\subsection{Ablation study and Supervised benchmark on Yahoo dataset}
\label{appendix:supervised_yahoo}

Here we present the results of our method on the supervised Yahoo dataset. It is important to note that, since the only baseline method that we found evaluated their model with point-wise F1 score, this is also what we use here to make our results comparable.

\begin{table}[tbh]
    \caption{Supervised anomaly detection performance on \textsc{Yahoo}. Results for \textsc{U-Net} taken from \citep{Gao2020robustad}.}
    \label{table:benchmark_yahoo_supervised}
    \begin{center}
        \begin{sc}
            \footnotesize{}
            \renewcommand{\arraystretch}{1.1}
                \begin{tabular}{lccccccccccccc}
                \toprule
                    & \multicolumn{3}{c}{Yahoo} \\ \cmidrule{2-4}
                    Model & F1 & prec & rec  \\
                \midrule
                    U-Net-Raw & 40.3 & 47.3  & 35.1 \\
                    U-Net-De & 62.1 & 65.1  & 59.4 \\
                    U-Net-DeW   & 66.2 & 79.3  & 56.9 \\
                    U-Net-DeWA  & 69.3 & 85.9  & 58.1 \\

                    
                    \midrule
                    \multicolumn{1}{l}{\acrshort{ncad} supervised} & 62.11 &  80.44  & 50.59 \\
                    \hspace{6pt} + mixup & 63.08 & 76.70  &  53.57  \\
                    \hspace{6pt} + \acrshort{po} & \textbf{79.92} & 74.96  &  85.57  \\
                    \hspace{6pt} + \acrshort{coe}  & 53.66 &  78.84  & 40.67 \\
                    \hspace{6pt} + \acrshort{coe} + mixup & 59.85 & 78.89  & 48.21 \\
                    \hspace{6pt} + \acrshort{po}\ + \acrshort{coe} & 58.36 &  54.89 & 62.30 \\
                    \hspace{6pt} + \acrshort{po}\ + \acrshort{coe} + mixup& 67.32 & 88.38  & 54.36 \\
                    \hspace{6pt} - contextual  & 5.50 & 3.42  & 14.08 \\
                    \hspace{6pt} - contextual + \acrshort{po}\  & 67.90 & 64.15  & 72.13 \\
                    \hspace{6pt} - contextual + \acrshort{coe} & 39.53 & 42.56   & 36.90 \\
                    \hspace{6pt} - contextual + \acrshort{po}\  + \acrshort{coe}   & 55.25  & 43.87  & 74.60  \\
                    \bottomrule
                \end{tabular}
        \end{sc}
    \end{center}
\end{table}

The supervised approach proposed by \citet{Gao2020robustad} is based on a U-net architecture, which is combined with preprocessing (using robust time series decomposition), loss weighting (to up-weight the rare anomalous class), and several forms of tailored data augmentation applied to the time series (keeping the labels unchanged). They report results for four variants: \textsc{U-Net-Raw} (plain supervised U-net on raw data), \textsc{U-Net-De} (applied to residual after preprocessing), \textsc{U-Net-DeW} (with loss weighting), \textsc{U-Net-DeWA} (with loss weighting and data augmentation).

Using only the true labels but no data augmentation (\textsc{\acrshort{ncad} supervised}), our approach significantly outperforms \textsc{U-Net-Raw}, and performs on-par with \textsc{U-Net-De}, without relying on time series decomposition and using an arguably much simpler architecture.

When we use the \acrshort{po}\ data augmentation, our approach outperforms the full \textsc{U-Net-DeWA} by a large margin, hinting at the possibility that addressing the class imbalance problem by creating artificial anomalies is more effective than using their strategy of loss weighting while keeping the labels intact.

In the supervised setting, injecting the generic \acrshort{coe} anomalies (either individually or in combination with \acrshort{po}) hurts performance, presumably by steering the model away from the specific kind of anomalies that are labeled as anomalous in this data set. On the other hand, adding \textsc{mixup} generally improves performance. The contrastive loss is crucial for good performance, as shown by the rows labeled \mbox{\textsc{- contrastive}}, where it is replaced with a standard softmax classifier.

\section{Model Implementation and Training}
\label{appendix:ncad_implementation}

At the core of the \acrshort{ncad} framework, we use a \emph{single} Encoder, $\phi(\cdot; \theta)$, to produce time series representations. The same encoder is applicable to all the time series in a given dataset, and it is used to encode both the full windows and the context windows. The parameters of the encoder, $\theta$, are learned via Mini-Batch Gradient Descent, aimed at minimizing the classification loss discussed in \cref{sec:ncad_architect_loss}.

Training mini-batches of size $B$ are created by first randomly selecting $b_{s}$ series from the training set, and then taking $b_{c}$ random fixed-size windows from each \footnote{\texttt{num\_series\_in\_train\_batch} and \texttt{num\_crops\_per\_series} in the supplementary code}. Data augmentation strategies described in \cref{sec:data_augmentation} are applied to these windows, creating additional examples which are incorporated to the batch. The number of augmented examples is controlled as a proportion of the original batch, using two additional hyperparameters: $r_{coe}$ and $r_{mixup}$ for \acrshort{coe} and Mixup, respectively. The size of the training batch is therefore
\[
B = b_{s} b_{c} + \floor{b_{s} b_{c} r_{coe}} + \floor{ b_{s} b_{c} r_{mixup} }
\]

Our implementation\footnote{open-source code available at \texttt{Anonymous Github repository}} is based on PyTorch \citep{Paszke2019pytorch} and PyTorch Lightning \citep{falcon2019pytorch}.
We used the default initialization defined in PyTorch, and the \emph{YOGI} optimizer \citep{Zaheer2018yogi} for all our experiments. We use standard AWS EC2 ml.p3.2xlarge instances with a single-core Tesla V100 GPU. Training and hyperparameter tuning was aided by AWS Sagemaker \citep{Liberty2020sagemaker}, training takes on average 90 minutes for each dataset.

\subsection{Model Hyperparameters}
Hyperparameters in our framework con be mainly divided in four categories:
\begin{enumerate}
    \item \textbf{Encoder architecture}: Number of TCN layers, TCN kernel size, embedding dimension, embedding normalization.
    \item \textbf{Data augmentation}: $r_{coe}$, $r_{mixup}$ (described above).
    \item \textbf{Optimizer}: learning rate, number of epochs.
    \item \textbf{Mini-batch cropping}: window length, suspect window length, $b_{s}$, $b_{c}$.
\end{enumerate}

For \textsc{Yahoo}, \textsc{KPI} and \acrshort{swat}, validation labels are available, so we use a Bayesian optimization \citep{perrone2020smtuner} for hyperparameter tuning, maximizing the F1 score on the validation set. We restricted the search of hyperparameters to only ``sensible'' values for most of the hyperparameters (e.g. max. 10 TCN layers, max. 256 dimensions for the embedding, max. 2.0 for the augmentation rates, etc.). Lengths of the window and suspect window are set by observing the lengths and seasonal patterns of the training dataset, so that it covers at least one cycle and this seasonality could be encoded in the representation. We use early stopping and keep the model with the lower validation F1, which is then evaluated on the test dataset and the result is reported.

For \acrshort{smap}, \acrshort{msl} and \acrshort{smd}, we do not have validation data to pick the hyperparametes, so we use default values that seemed to work well for the other datasets. It is not possible to do early stopping either, so we keep the model resulting of training until the last epoch, which is then evaluated on the test dataset and the result is reported.

In all cases, we align our experimental protocol with prior works and report $F1$ scores computed by choosing the best threshold on the test set.

We provide hyperparameter configuration files in the supplementary code, which allow to replicate our benchmark results in section \cref{sec:experiments}.

\newpage

\section{Datasets and External Assets}
\label{appendix:data_and_assets}

We use the following datasets to compare the performance of \acrshort{ncad} to other methods:

\textbf{\acrfull{smap}} and \textbf{\acrfull{msl}} \cite{Hundman2018telemanon}, the datasets are under the custom license \url{https://github.com/khundman/telemanom/blob/master/LICENSE.txt}.

\textbf{\acrfull{swat}} \citep{mathur2016swat} This dataset is distributed by the ITrust Centre for Research in Cyber Security \url{https://itrust.sutd.edu.sg/itrust-labs_datasets/dataset_info/}, we were not able to find the precise license of the dataset.

\textbf{\acrfull{smd}} \citep{omnianomaly} is distributed under the MIT License \url{https://github.com/NetManAIOps/OmniAnomaly/blob/master/LICENSE}.

\textbf{\textsc{Yahoo}} A dataset published by Yahoo labs, \url{https://webscope.sandbox.yahoo.com/catalog.php?datatype=s&did=70}, we were not able to find the precise license of the datase beyond the ReadMe.txt specifying that the dataset could be used for non-commercial research purposes.

\textbf{\textsc{KPI}} \citep{kpi1} we were not able to find the precise license of the dataset.

\paragraph{Additional assets}
In addition to the datasets, we use existing code for the TCN encoder from \url{https://github.com/White-Link/UnsupervisedScalableRepresentationLearningTimeSeries} which is under the Apache License Version 2.0.
We also use the evaluation code from \url{https://github.com/NetManAIOps/OmniAnomaly/blob/master/donut_anomaly/eval_methods.py} which is under the MIT License.
We use the standard Python library numpy \cite{harris2020array}, which is under the BSD 3-Clause "New" or "Revised" License \url{https://github.com/numpy/numpy/blob/main/LICENSE.txt}.

We make our code available, licensed under the Apache License, Version 2.0.

All the dataset and code that we use in this work is openly available under licences that allow to use them, as a result we did not seek additional consent from their creators.
None of the datasets contains personally identifiable information, nor do they contain offensive content.

\clearpage
\printglossaries

\end{document}